\documentclass[11pt]{article}
\usepackage[final]{acl}
\usepackage{times}
\usepackage{latexsym}
\usepackage[T1]{fontenc}
\usepackage[utf8]{inputenc}
\usepackage{microtype}
\usepackage{inconsolata}
\usepackage{graphicx}
\usepackage[table]{xcolor}
\usepackage{multirow}
\usepackage{amssymb} 
\usepackage{booktabs}
\usepackage{adjustbox}
\usepackage{enumitem}
\usepackage{stfloats}
\usepackage[most]{tcolorbox}
\usepackage{tabularx}
\usepackage{ragged2e}
\usepackage{array}
\usepackage{hyperref}
\usepackage{xcolor}
\usepackage{mdframed}
\usepackage{caption}
\definecolor{navyblue}{RGB}{23,54,93} 
\definecolor{QAFrame}{RGB}{215,215,215}
\definecolor{CardBg}{RGB}{252,252,252}
\definecolor{TitleBg}{RGB}{238,238,238}
\definecolor{QBg}{RGB}{242,246,255}
\definecolor{ABg}{RGB}{243,251,245}
\definecolor{TextDark}{RGB}{45,45,45}

\newtcolorbox{qacard}[1][]{
  enhanced,
  colback=CardBg,
  colframe=QAFrame,
  boxrule=0.6pt,
  arc=3mm,
  left=2.2mm,right=2.2mm,top=2.0mm,bottom=2.0mm,
  before=\par\noindent\begin{samepage},
  after=\end{samepage}\par,
  clip upper,
  clip lower,
  #1
}

\title{Can MLLMs Reason Beyond Language? \\
VisReason: A Comprehensive Benchmark for Vision-Centric Reasoning}

\author{
Longteng Guo\textsuperscript{1,2}\thanks{Equal contribution.}\qquad
Yifan Wang\textsuperscript{1,2}\footnotemark[1]\qquad
Pengkang Huo\textsuperscript{1,2}\footnotemark[1]\qquad
Tailai Chen\textsuperscript{1,2} \\
\textbf{Yuze Wu}\textsuperscript{1,2}\qquad
\textbf{Jing Liu}\textsuperscript{1,2}\qquad
\textbf{Xinxin Zhu}\textsuperscript{1,2}\thanks{Corresponding authors.} \\
\textsuperscript{1}Institute of Automation, Chinese Academy of Sciences \\
\textsuperscript{2}School of Artificial Intelligence, University of Chinese Academy of Sciences \\
\texttt{\{longteng.guo, jliu, xinxin.zhu\}@nlpr.ia.ac.cn} \\
\texttt{\{wangyifan2026, huopengkang2025\}@ia.ac.cn}
}

\begin{document}
\maketitle


\begin{abstract}
Recent multimodal large language models (MLLMs) achieve strong performance on visual reasoning benchmarks, yet it remains unclear to what extent such performance reflects reasoning directly grounded in visual evidence. We introduce VisReason, a benchmark for vision-centric reasoning in everyday scenarios where perception and inference are tightly coupled. VisReason contains 1,505 questions across 10 categories spanning perceptual, structural, and conceptual reasoning. Our evaluation shows that VisReason poses a qualitatively different challenge from existing benchmarks, exposing substantial gaps between humans and current MLLMs and revealing limited benefits from test-time reasoning strategies. VisReason offers a focused diagnostic for evaluating vision-centric reasoning beyond language. We release our resources at \url{https://github.com/CASIA-IVA-Lab/VisReason}
\end{abstract}

\begin{figure*}[t]
    \centering
    \includegraphics[width=\textwidth]{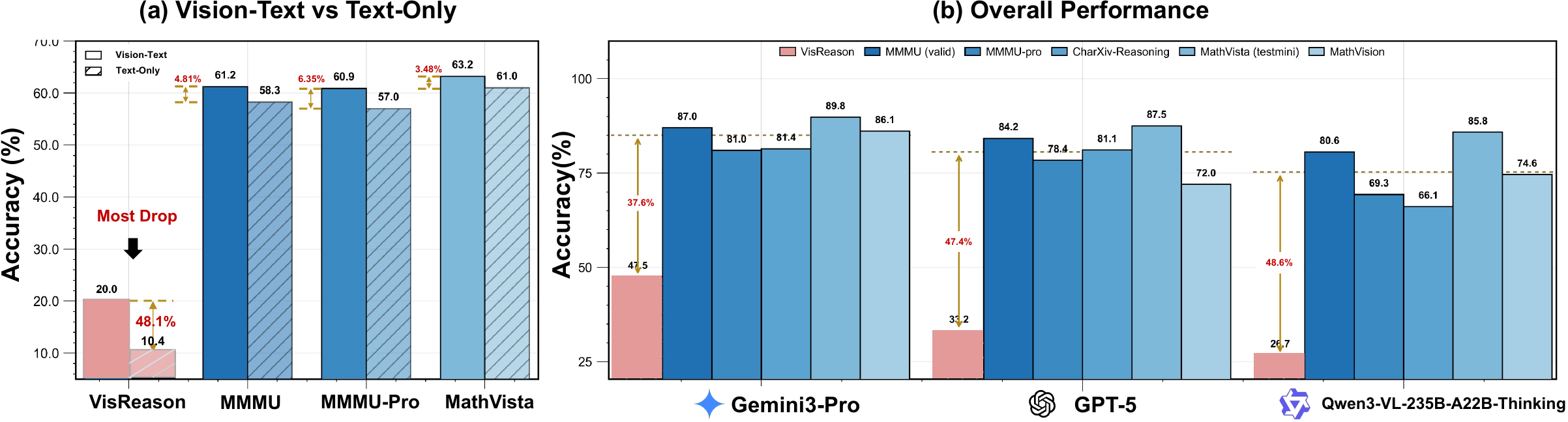}
    \caption{
    Diagnostic comparison of vision-centric reasoning across benchmarks and models.
    \textbf{Left:} Performance under direct vision--text inputs versus language-mediated text-only inputs, evaluated using the same MLLM (Qwen3-VL-32B), where images are replaced by captions generated by Qwen-VL-Max, highlighting the reliance on visual evidence.
    \textbf{Right:} Overall performance of representative proprietary and open-source MLLMs on VisReason and existing visual reasoning benchmarks, showing a larger performance gap on VisReason.
    }
    \label{fig:img-Diagnostic}
\end{figure*}


\section{Introduction}
\label{sec:intro}
Recent advances in multimodal large language models (MLLMs) demonstrate strong performance on tasks such as scientific question answering, diagram understanding, and mathematical problem solving, suggesting an emerging ability to integrate visual and textual information for non-trivial reasoning.

However, many widely used visual reasoning benchmarks emphasize STEM-oriented or knowledge-intensive problems, where textual descriptions, symbolic representations, or domain priors play a dominant role. In these settings, models can often abstract visual inputs into language and perform most reasoning in the linguistic space, raising a fundamental question: \emph{do current MLLMs truly reason from visual evidence, or do they primarily rely on language-mediated inference?}

To probe this question, we conduct a diagnostic comparison across existing benchmarks, as shown in Fig.~\ref{fig:img-Diagnostic}. Using the same MLLM, we compare performance under direct vision--text inputs against a language-mediated setting where images are replaced by model-generated captions. On established benchmarks such as MMMU \cite{Yue_2024_CVPR}, MMMU-Pro \cite{yue-etal-2025-mmmu}, and MathVista \cite{lu2024mathvista}, this substitution leads to only minor performance drops (< 6.35\%), indicating that much of the reasoning can be preserved through textual descriptions alone. 

These observations suggest that current evaluation protocols provide limited insight into vision-centric reasoning, as models can often succeed through language-mediated abstractions without robust visual grounding. By contrast, human visual reasoning routinely operates directly on visual input, enabling people to infer structure, relations, and implicit rules without relying on rich language descriptions or formal knowledge. This contrast raises a central question: \textit{can MLLMs reason in a similarly vision-centric manner?}

To answer this question, we introduce \textbf{VisReason}, a benchmark designed to evaluate vision-centric reasoning beyond language mediation. VisReason targets everyday visual scenarios in which perception and reasoning are tightly coupled and textual cues alone are insufficient. As shown in Fig.~\ref{fig:img-Diagnostic}, removing visual input leads to a substantial performance drop (48.12\%), underscoring VisReason’s reliance on direct visual evidence.

VisReason consists of 1,505 carefully curated questions spanning 10 reasoning categories, covering perceptual, structural, and conceptual forms of visual reasoning, such as identifying visual differences, reasoning about 3D-spatial and game-board configurations, and inferring implicit rules from visual cues.

By isolating and evaluating vision-centric reasoning across these dimensions, VisReason offers a diagnostic lens on the limitations of current MLLMs and a foundation for advancing multimodal models that reason more directly from visual evidence. Our experiments on VisReason lead to several key findings.

\begin{itemize}[leftmargin=*, itemsep=2pt]
\item \textbf{VisReason exposes a large and diagnostic gap between humans and leading MLLMs.}
Humans solve VisReason reliably using the provided visual evidence, whereas even the strongest MLLMs remain far behind.
Moreover, VisReason sharply differentiates model capabilities, with performance ranging from near-chance to substantially higher accuracy.

\item \textbf{Increasing inference-time thinking budget yields consistent but slow and saturating gains.}
Allocating more inference-time tokens generally improves accuracy, but the gains grow gradually and tend to plateau at higher budgets.
As a result, additional test-time computation offers limited leverage in the absence of stronger visual grounding.

\item \textbf{Explicit CoT prompting provides only marginal benefit for non-thinking models.}
For non-thinking models, explicit CoT prompting yields only small average improvements (e.g., +1.1\% for GPT-4o) and does not consistently benefit perception-heavy categories.
Without accurate extraction of visual evidence, longer textual reasoning traces rarely translate into better answers.

\item \textbf{Scaling model capacity consistently improves vision-centric reasoning performance.}
Across model families, larger models achieve higher accuracy on VisReason, confirming that model capacity remains an important contributor to vision-centric reasoning.
Nevertheless, even the largest models fall well short of human performance, leaving substantial room beyond scaling alone.
\end{itemize}

\begin{figure*}[!t]
    \centering
    \includegraphics[width=\textwidth]{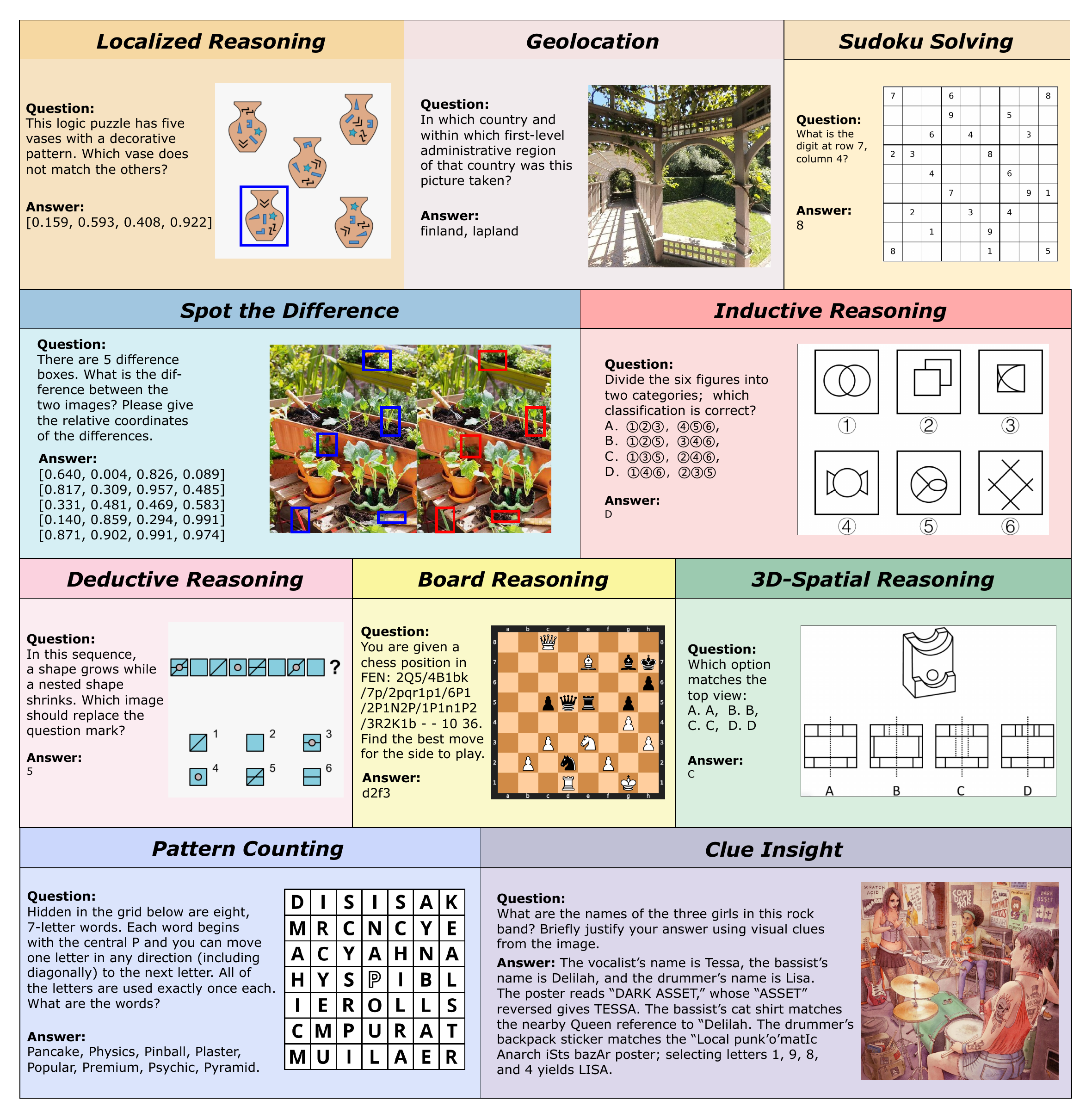}
    \vspace{-0.2cm}
    \caption{VisReason examples from different reasoning categories, covering everyday visual reasoning scenarios where perception and reasoning are tightly coupled.}
    \label{fig:img-DataDemos}
\end{figure*}


\section{Related Work}

\subsection{Multimodal Large Language Models}
Recent MLLMs have shown strong multimodal understanding and are increasingly tested on whether they can carry out multi-step reasoning grounded in visual evidence.
Representative proprietary systems include GPT-4V \cite{openai2023gpt4v}, GPT-4o \cite{openai2024gpt04o}, Gemini \cite{team2023gemini}, and Claude, while competitive open-weight families include Qwen-VL \cite{wang2024qwen2} and InternVL \cite{chen2024internvl,zhu2025internvl3}.
In addition, thinking-style variants \cite{guo2025deepseek} allocate larger reasoning budgets at inference time, motivating closer examination of whether the gains come from stronger vision-centric reasoning with tighter visual grounding, rather than merely longer language-side deliberation.

\subsection{Visual Reasoning Benchmarks}
Recent multimodal evaluation has increasingly emphasized reasoning over visual evidence.
Benchmarks such as ScienceQA \cite{lu2022learn}, MathVista \cite{lu2024mathvista}, and MMMU \cite{Yue_2024_CVPR} introduce exam-style, visually grounded problems, but remain largely STEM- and knowledge-centered and do not always require fine-grained, directly verifiable visual evidence.
In parallel, domain-specific benchmarks such as Sudoku-Bench \cite{seely2025sudokubenchevaluatingcreativereasoning} and puzzle-style datasets like REBUS \cite{gritsevskiy2024rebus} and VGRP-Bench \cite{ren2025vgrp-bench} focus on narrower aspects of vision-centric reasoning, typically constrained to games, symbols, or rule-based settings. 
VisReason complements prior work by targeting common-context vision-centric reasoning across a broader spectrum of visual scenarios. 
It systematically covers perceptual, structural, and conceptual reasoning behaviors, organizing tasks into 10 categories and 36 subcategories to enable more comprehensive evaluation.


\section{The VisReason Benchmark}

\subsection{Overview of VisReason}
\begin{figure*}[!h]
    \centering
    \includegraphics[width=\textwidth]{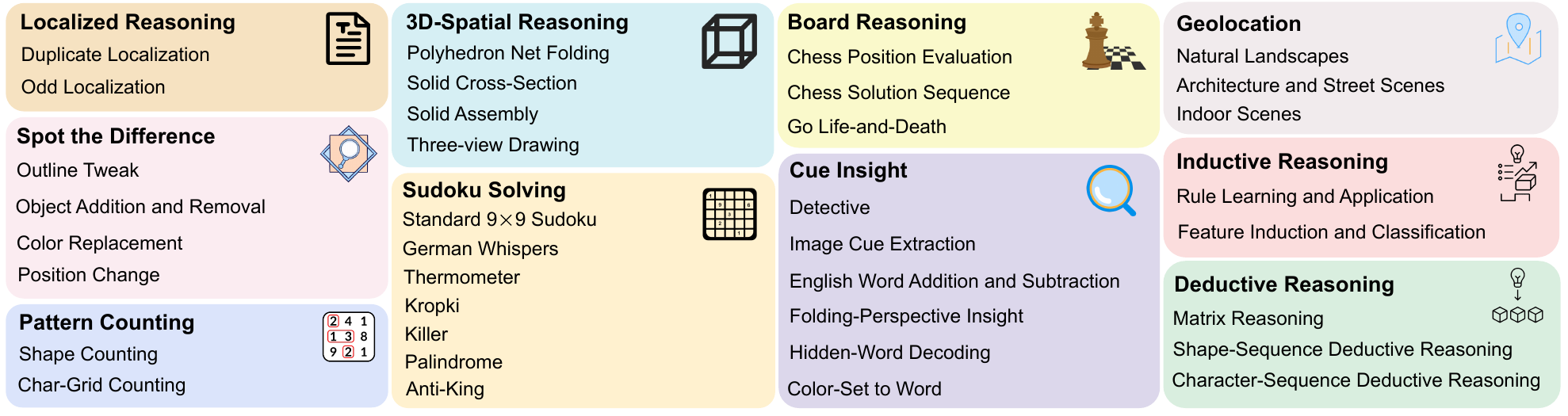}
    \caption{Domain diversity in VisReason. The benchmark covers 10 reasoning categories and 36 fine-grained subcategories, illustrating the breadth of vision-centric reasoning tasks.}
    \label{fig:img-DataAll}
\end{figure*}

We propose VisReason, a multimodal benchmark designed to evaluate foundation models on vision-centric reasoning in everyday, perceptually grounded contexts. VisReason systematically spans perceptual, structural, and conceptual reasoning, covering abilities from low-level visual grounding to higher-level abstraction and inference. As illustrated in Figure~\ref{fig:img-DataAll}, which summarizes the category structure, the benchmark comprises 10 reasoning categories organized into 36 fine-grained subcategories. Figure~\ref{fig:img-DataDemos} further provides representative examples from each category. Together, these categories capture a broad spectrum of visual reasoning phenomena and emphasize tasks that require direct reasoning over visual evidence, rather than relying primarily on language priors or knowledge-intensive inference.

\subsection{Data Curation Process}

\noindent \textbf{Data Collection.} We first compile a candidate set of 10 visual reasoning categories. 
Guided by this candidate set, we curate questions from online sites and Chinese civil service exam repository. 
During data collection, we explicitly prioritized vision-centric items whose solutions hinge on extracting and integrating visual evidence, with a particular focus on graphic/diagrammatic reasoning problems that require interpreting shapes, layouts, and fine-grained visual cues rather than relying on text-only inference. During data collection, we manually verified that the selected samples do not contain personally identifying information.

\noindent \textbf{Question Revision.} After collecting the data, we recruited graduate students to convert the raw samples into a standardized visual question answering format.
Firstly, we categorize questions by answer type into four forms: multiple-choice question, fill in the blank, open ended solution, and 2D bounding box selection. 
Each sample is independently judged by at least three annotators, and the final type is determined by majority vote.
Secondly, we standardize questions and answers by format. 
For multiple-choice, we ensure the question contains explicit option indices and texts, and we restrict the answer to the option index only. 
For fill-in-the-blank question, we remove any extraneous content from the answer, keeping only the filled text. 
For open-ended, we ensure the question poses a clear, specific query.
For bounding-box, annotators manually mark the target instance in the image and convert it to the [x1, y1, x2, y2] format as the final answer.\\

\noindent \textbf{Manual Validation.} 
To ensure data quality, we conducted rigorous manual validation for all entries. Each image–question–answer triplet was independently reviewed by two annotators, who verified basic correctness and filtered out problematic cases. Disagreements were resolved through discussion and adjudication. 
During review, annotators evaluated multiple aspects of data quality.
(1) Completeness and consistency: We checked that the image, question, and answer are mutually consistent and that no essential information is missing.
(2) Visual evidence sufficiency: Annotators assessed whether the key visual cues required to solve the question are clearly observable in the image and support a unique, well-defined answer.
(3) Answer clarity: For multiple-choice questions, we verified that all candidate options are explicitly specified in the question text or visually present in the image, ensuring that each question is self-contained and the correct option is unambiguous.
(4) Multimodal necessity and knowledge scope: We removed items that could be answered using only visual information or only textual information. In addition, we filtered out questions that depend on external knowledge beyond the provided context, so that solving each item relies primarily on reasoning over the given visual and textual evidence.


\subsection{Reasoning Category}
Each question in VisReason is manually assigned to a specific reasoning category.
The 10 categories are organized under the three-level framework of \emph{perceptual}, \emph{structural}, and \emph{conceptual} reasoning, spanning from low-level visual grounding to high-level abstraction and inference. Together, they provide a structured view of vision-centric reasoning behaviors grounded in visual evidence and commonly encountered in real-world scenarios.

\begin{itemize}[itemsep=0pt, parsep=0pt, topsep=0pt]
\item \textbf{Localized Reasoning}: Inferring the target instance and localizing it with a bounding box.
\item \textbf{Spot the Difference}: Identifying visual differences between two scenes and localizing each difference with bounding boxes.
\item \textbf{Pattern Counting}: Identifying all target patterns in complex scenes and counting their occurrences.
\item \textbf{3D-Spatial Reasoning}: Reasoning about implicit 3D geometric relationships to infer the correct answer.
\item \textbf{Board Reasoning}: Inferring the correct outcome given a board-game configuration.
\item \textbf{Sudoku Solving}: Solving multiple variants of visually grounded Sudoku puzzles.
\item \textbf{Geolocation}: Extracting geographic cues from the visual scene to infer its location.
\item \textbf{Cue Insight}: Extracting and integrating textual and visual clues to derive the answer.
\item \textbf{Inductive Reasoning}: Inferring underlying rules from observed instances and applying them to novel cases.
\item \textbf{Deductive Reasoning}: Deriving logically entailed conclusions from the given visual and contextual information.
\end{itemize}

\begin{table}[t]
  \caption{Key statistics of VisReason.}
  \centering
  \small
  \begin{tabular}{lr}
    \toprule
    \textbf{Statistic} & \textbf{Number}\\
    \midrule
    \#Total questions & 1505\\
    - Multiple-choice & 615 (40.9\%)\\
    - Fill-in-the-blank & 441 (29.3\%)\\
    - Open-ended & 309 (20.5\%)\\
    - Bounding-box & 140 (9.3\%)\\
    
    \midrule
    \#Reasoning categories & 10 \\
    \#Subcategories & 36 \\
    \midrule
    Question length (min / avg / max) & 5 / 45.9 / 282 \\
    Answer length (min / avg / max) & 1 / 9.1 / 144 \\
  \bottomrule
\end{tabular}
\label{tab-statics}
\end{table}


\subsection{Dataset Statistics}

Table~\ref{tab-statics} summarizes the key statistics of VisReason.
The benchmark contains 1,505 questions designed to support systematic evaluation of vision-centric reasoning across diverse settings.

\indent VisReason is organized into 10 reasoning categories that span multiple levels of visual cognition, from perceptual grounding to abstract inference.
Building on these core categories, the dataset further covers 36 fine-grained subcategories, capturing a broad spectrum of visual reasoning phenomena.
Representative examples from each category and subcategory are provided in the Appendix.

\indent VisReason includes four question formats that differ in how answers are specified and constrained.
Multiple-choice, fill-in-the-blank, and open-ended questions impose different levels of answer openness, 
while all requiring reasoning over the visual input.
In addition, VisReason contains a set of bounding-box questions arising from the Localized Reasoning and Spot-the-Difference categories,
where models are required to infer the correct answer and identify its spatial location within the image.
This format explicitly involves spatial localization as part of the reasoning process.

\indent Overall, the dataset exhibits substantial linguistic and contextual diversity.
Question lengths range from short prompts to complex multi-sentence descriptions, with an average length of 45.9 tokens.

\begingroup
\setlength{\tabcolsep}{4.8pt} 
\begin{table*}[t]
    \centering
    \small
    \caption{Comparison of model performances across reasoning categories on VisReason. \colorbox{navyblue!5}{Gray}  rows indicate non-thinking models, while others employ explicit reasoning mechanisms.}
    \label{tab:overall-results}
    \begin{tabular}{@{}l|r|rrrrrrrrrr@{}}
        \toprule
        & &
        \multicolumn{1}{c}{\rotatebox{75}{\textbf{Loc. Reas.}}} &
        \multicolumn{1}{c}{\rotatebox{75}{\textbf{Spot Diff.}}} &
        \multicolumn{1}{c}{\rotatebox{75}{\textbf{Pattern Count.}}} &
        \multicolumn{1}{c}{\rotatebox{75}{\textbf{3D-Spat. Reas.}}} &
        \multicolumn{1}{c}{\rotatebox{75}{\textbf{Board Reas.}}} &
        \multicolumn{1}{c}{\rotatebox{75}{\textbf{Sudoku Solv.}}} &
        \multicolumn{1}{c}{\rotatebox{75}{\textbf{Geolocation}}} &
        \multicolumn{1}{c}{\rotatebox{75}{\textbf{Cue Insight}}} &
        \multicolumn{1}{c}{\rotatebox{75}{\textbf{Ind. Reas.}}} &
        \multicolumn{1}{c}{\rotatebox{75}{\textbf{Ded. Reas.}}} \\
        \textbf{Models} & \multicolumn{1}{c}{\textbf{Avg.}} &
        \multicolumn{3}{c}{\cellcolor{orange!15}\textit{Perceptual}} &
        \multicolumn{3}{c}{\cellcolor{yellow!12}\textit{Structural}} &
        \multicolumn{4}{c}{\cellcolor{green!10}\textit{Conceptual}} \\
        \midrule

        Human Performance         & 71.4 & 81.7 & 77.4 & 66.9 & 71.6 & 56.9 & 45.6 & 80.0 & 74.8 & 80.6 & 78.1 \\
        \midrule
         \multicolumn{12}{c}{\textbf{\textit{Proprietary Models}}} \\
        \rowcolor{navyblue!5} GPT-4o             & 15.6 &  6.3 &  0.8 &  8.7 & 25.4 & 18.5 &  7.5 &  5.5 & 27.1 & 23.6 & 32.1 \\
        o4-mini            & 27.3 &  3.8 &  2.7 & 21.7 & 27.7 & 26.7 & 15.5 & 55.5 & 46.0 & 36.4 & 36.9 \\
        GPT-5-nano         & 19.0 &  3.8 &  0.1 & 17.4 & 23.9 & 22.2 &  9.0 & 29.0 & 29.7 & 29.8 & 24.6 \\
        GPT-5-mini (low)   & 28.1 & 12.5 &  2.2 & 15.2 & 30.8 & 30.4 & 15.0 & 55.0 & 45.1 & 37.5 & 37.7 \\
        GPT-5-mini (medium)& 31.8 &  7.5 &  2.0 & 32.6 & 28.5 & 36.3 & 12.5 & 56.5 & 55.9 & 41.8 & 44.4 \\
        GPT-5-mini (high)  & 33.2 & 12.5 &  1.1 & 37.0 & 26.9 & 42.2 &  8.0 & 61.0 & 60.4 & 41.8 & 41.4 \\
        GPT-5              & 33.2 & 15.0 &  2.8 & 21.7 & 26.2 & 44.4 & 22.5 & 66.5 & 53.2 & 40.7 & 38.8 \\
        GPT-5.2            & 35.5 & 24.6 &  4.7 & 26.1 & 30.8 & 43.7 & 38.2 & 52.5 & 43.2 & 42.3 & 48.9 \\
        Gemini-2.5-Flash   & 16.0 &  0.0 &  0.5 &  8.7 & 31.5 & 15.6 &  6.0 & 12.5 & 19.8 & 31.6 & 34.0 \\
        Gemini-3-Pro       & 47.5 & 44.2 &  4.8 & 50.0 & 43.9 & 51.1 & 30.0 & 76.0 & 68.5 & 53.1 & 53.7 \\
        \midrule
        \multicolumn{12}{c}{\textbf{\textit{Open-source Models}}} \\
        \rowcolor{navyblue!5} InternVL3-8B                 & 10.9 &  0.0 &  0.1 &  2.2 & 31.5 &  6.7 &  8.5 & 11.0 &  9.9 & 19.3 & 20.2 \\
        \rowcolor{navyblue!5} InternVL3-14B                & 10.4 &  1.3 &  0.1 &  6.5 & 22.3 &  5.9 &  2.5 & 14.0 & 12.6 & 20.0 & 19.0 \\
        \rowcolor{navyblue!5} Qwen2.5-VL-7B-Instruct       &  7.2 &  0.0 &  0.3 &  0.0 & 18.5 &  1.5 &  1.5 &  0.0 &  9.0 & 22.2 & 19.0 \\
        \rowcolor{navyblue!5} Qwen2.5-VL-32B-Instruct      & 15.4 &  3.3 &  1.5 &  6.5 & 27.7 &  8.2 &  8.0 & 29.0 & 21.6 & 25.1 & 23.5 \\
         \rowcolor{navyblue!5} Qwen3-VL-8B-Instruct         & 18.8 &  1.3 &  1.1 &  8.6 & 22.8 & 13.1 &  9.2 & 29.5 & 34.3 & 33.7 & 34.1 \\
        Keye-VL-1.5-8B               & 10.9 &  0.0 &  0.8 & 10.9 & 19.2 &  6.7 &  4.5 &  5.0 & 16.6 & 24.9 & 20.5 \\
        MiMo-VL-7B-RL                & 15.9 &  2.1 &  0.7 & 10.6 & 25.4 & 17.0 &  2.5 & 37.5 & 12.6 & 24.7 & 25.4 \\
        Qwen3-VL-2B-Thinking         &  3.7 &  2.5 &  0.2 &  2.2 &  3.9 &  2.2 &  4.0 &  3.0 &  5.4 &  8.0 &  6.0 \\
        Qwen3-VL-8B-Thinking         & 19.0 &  9.6 &  0.7 &  6.5 & 33.1 & 14.1 &  6.0 & 25.3 & 25.2 & 35.0 & 34.0 \\
        Qwen3-VL-30B-A3B-Thinking    & 20.0 &  0.0 &  0.5 & 15.2 & 30.8 & 17.0 &  8.5 & 22.5 & 28.4 & 34.9 & 37.3 \\
        Qwen3-VL-32B-Thinking        & 22.1 &  1.3 &  1.3 & 17.4 & 30.8 & 27.4 &  7.0 & 31.0 & 27.0 & 38.6 & 39.2 \\
        Qwen3-VL-235B-A22B-Thinking  & 26.7 &  7.1 &  1.6 & 19.6 & 36.2 & 30.4 &  4.0 & 46.0 & 30.6 & 46.2 & 45.2 \\
        \bottomrule
    \end{tabular}
\end{table*}
\endgroup


\section{Experiments}

\subsection{Experimental Setup}
\paragraph{Metrics.} 
We report accuracy ($\%$) for all question types in VisReason, with evaluation protocols adapted to the corresponding answer formats.
For multiple-choice and fill-in-the-blank questions, the final answer is extracted using regular expressions and directly compared with the ground truth.
For open-ended questions, we employ GPT-5-mini as an automatic judge to assess correctness given the question and reference answer under a strict criterion; the full judging prompt is provided in the Appendix. 
For bounding-box questions, a single question may involve multiple ground-truth boxes.
Predicted boxes are matched to ground truth using IoU-based Hungarian matching, and a match is considered correct if IoU $>$ 0.5. This threshold was selected based on preliminary experiments, as it produces a more reasonable distribution of model performance and avoids settings that are either too loose or too strict.
Accuracy is computed with partial credit and normalized to 1 per question.

\noindent\textbf{Baseline Models.} 
We evaluate VisReason on a diverse set of proprietary and open-source MLLMs, covering a wide range of model architectures, parameter scales, and reasoning capabilities.

\textit{Proprietary Models.} We include (1) GPT-5.2,\allowbreak GPT-5, GPT-5-mini, o4-mini;\allowbreak (2) Gemini-2.5-Flash,\allowbreak Gemini-3-Pro \allowbreak.
For the flagship GPT-5 and Gemini-3 models, their reasoning effort is set to \textit{low} for cost control and evaluation consistency.

\textit{Open-source Models.} We evaluate a broad collection of leading open models, including (1) Qwen-VL series, including Qwen2.5-VL \cite{bai2025qwen2} (7B-Instruct, 32B-Instruct), Qwen3-VL \cite{bai2025qwen30vl}(8B-Instruct, 2B-Thinking, 8B-Thinking, 32B-Thinking, 30B-A3B-Thinking, 235B-A22B-Thinking);
(2) InternVL3 \cite{zhu2025internvl3} (8B, 14B); (3) MiMo-VL-7B-RL \cite{coreteam2025mimovltechnicalreport}; (4) Keye-VL-1.5-8B \cite{yang2025kwaikeyevl15technical}.

All models are evaluated in a zero-shot setting.
We design four prompt templates corresponding to the four question formats, all following a Chain-of-Thought \cite{wei2022chain} (CoT) prompting style.
The complete prompt templates are provided in the Appendix.

\textit{Human Performance.} We additionally include human performance to contextualize model results and to verify that VisReason tasks are reliably solvable given the provided visual evidence; the human evaluation protocol is detailed in the Appendix.


\subsection{Overall Results}

VisReason reveals a substantial gap between current MLLMs and human-level vision-centric reasoning.
Table~\ref{tab:overall-results} reports model performance across all reasoning categories. Overall accuracy remains low across the board: the best-performing model achieves only 47.5\% average accuracy, while many models fall below 20\%. In contrast, human performance reaches 71.4\%, indicating that VisReason is reliably solvable by humans under the provided visual evidence but remains highly challenging for current models. This gap confirms that VisReason effectively targets vision-centric reasoning abilities that are not yet well captured by existing MLLMs.

\noindent\textbf{Proprietary vs. Open-Source Models.}
State-of-the-art vision-centric reasoning is still dominated by proprietary models.
A clear performance stratification emerges between proprietary and open-source systems. The strongest proprietary model, Gemini-3-Pro, achieves an overall accuracy of 47.5\%, whereas the best open-source model, Qwen3-VL-235B-A22B-Thinking, reaches only 26.7\%. This gap suggests that large-scale pretraining and stronger cross-modal alignment remain critical for effective vision-centric reasoning.

\noindent\textbf{VisReason Clearly Differentiates Model Capabilities.}
Performance varies widely across models, with several weaker systems performing near chance level. 
For instance, Qwen3-VL-2B-Thinking attains an overall accuracy of only 3.7\%, while a substantial gap separates such models from the strongest systems. 
This spread indicates that VisReason distinguishes model capabilities across a wide range of capacities.


\subsection{Where Do MLLMs Struggle in Vision-Centric Reasoning?}
Despite steady progress in multimodal modeling, VisReason reveals that current MLLMs struggle when reasoning must be tightly grounded in visual evidence. As shown in Table~\ref{tab:overall-results}, across reasoning subjects, performance varies in a systematic but non-uniform manner: tasks that can be solved using coarse visual cues or scene-level abstraction are handled better, whereas tasks requiring fine-grained perceptual grounding and spatial precision remain challenging for most models.

These limitations are most evident in perception-grounded tasks that explicitly involve spatial localization. On Localized Reasoning and Spot-the-Difference, average model accuracy is only 6.9\% and 1.3\%, respectively, indicating substantial difficulty across all evaluated systems. In contrast, human performance on the same categories reaches 81.7\% and 77.4\%, confirming that these tasks are well-defined and reliably solvable given the visual evidence. The large human–model gap highlights persistent challenges in aligning inferred reasoning outcomes with localized visual evidence and in validating perceptual hypotheses against specific image regions.

\begin{figure*}[!t]
    \centering
    \includegraphics[width=0.9\textwidth]{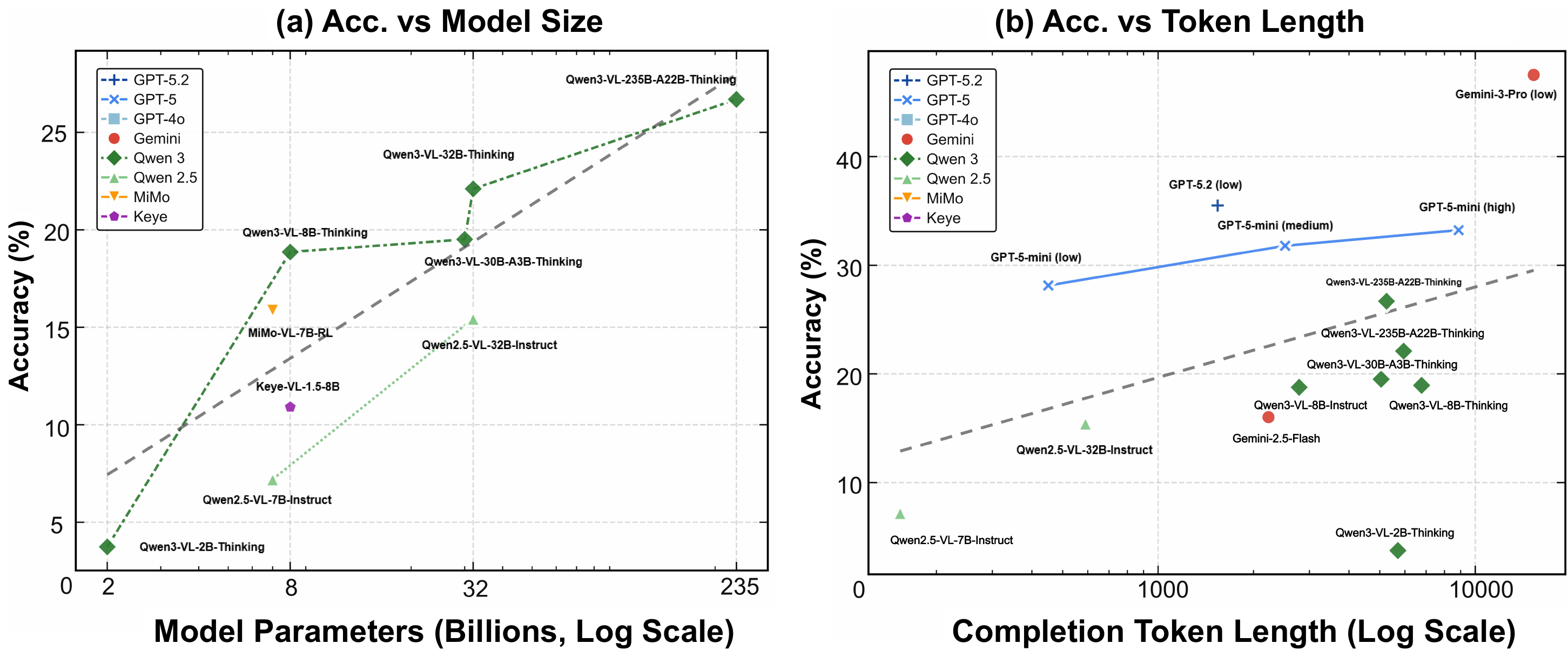}
    \caption{Scaling behavior of models on VisReason. \textbf{Left:} Accuracy versus model size (log scale). \textbf{Right:} Accuracy versus average inference-time tokens (log scale).}
    \label{fig:img-Cropped}
\end{figure*}

\begin{figure}
    \centering
    \includegraphics[width=\linewidth]{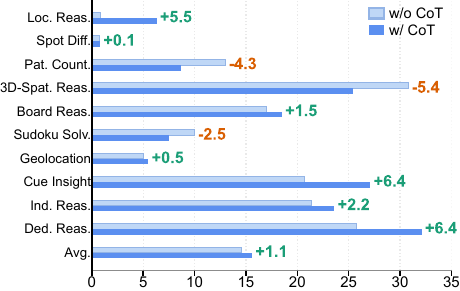}
    \caption{Effect of COT prompting on GPT-4o performance across different reasoning categories.}
    \label{fig:cot}
\end{figure}


\subsection{Does CoT Help in Answering Vision-Centric Questions?}
Figure~\ref{fig:cot} compares GPT-4o performance under two prompting settings: with and without CoT prompting. Overall, CoT yields only a marginal gain on VisReason, improving average accuracy by +1.1\%. This improvement is smaller than the commonly assumed effectiveness of CoT for reasoning. A category-wise breakdown shows that CoT mainly helps higher-level reasoning tasks, such as Cue Insight and inductive or deductive reasoning, while providing little benefit or even degrading performance on Spot-the-Difference, 3D-Spatial Reasoning, and Pattern Counting. These results suggest that longer reasoning traces alone are insufficient when task difficulty is dominated by visual perception and grounding.


\subsection{How Much Does Compute Scaling Improve Vision-Centric Reasoning?}
\noindent\textbf{Scaling Model Capacity.}
Figure~\ref{fig:img-Cropped} (left) reveals a clear scaling trend within model families: accuracy generally improves as model parameters increase. This effect is most evident along the upward trajectories of the Qwen2.5 and Qwen3 series, where larger variants consistently outperform their smaller counterparts. Overall, increasing model capacity provides a reliable improvement for vision-centric reasoning on VisReason.

\noindent\textbf{Increasing Inference-Time Thinking Budget.}
Figure~\ref{fig:img-Cropped} (right) shows that increasing the thinking budget leads to consistent but gradual accuracy improvements for the same model. This trend is most evident for GPT-5-mini, where moving from low to moderate token budgets yields noticeable gains, while further increases at higher budgets result in diminishing improvements, indicating that additional test-time computation offers limited leverage in the absence of stronger visual grounding.

\begin{figure}[t]
    \centering
    \includegraphics[width=\linewidth]{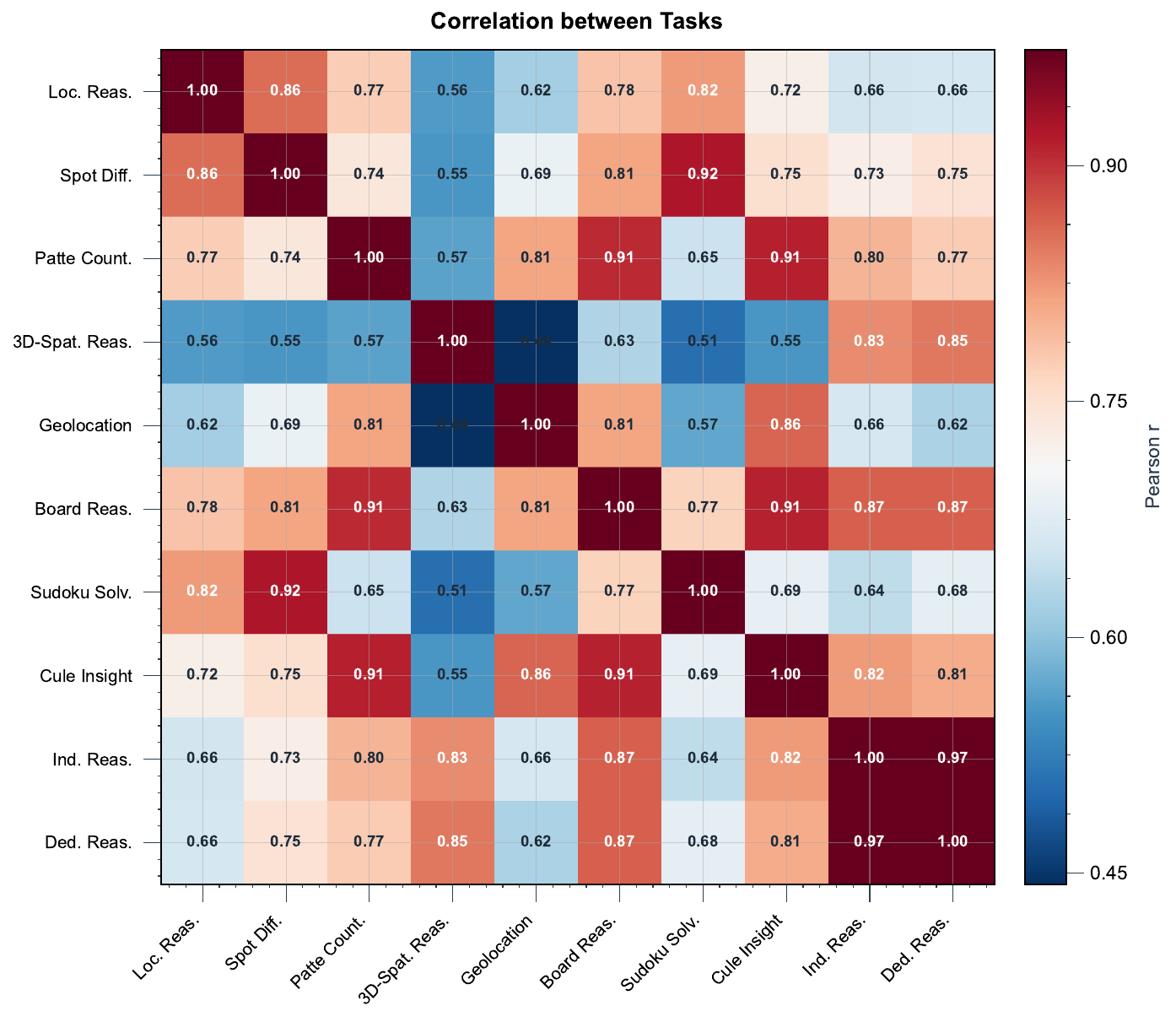}
    \caption{Pearson correlation of model accuracy across the 10 reasoning categories on VisReason.}
    \label{fig:heatmap}
\end{figure}


\subsection{Are Reasoning Abilities Correlated Across Categories?}
Figure~\ref{fig:heatmap} shows the Pearson correlation of model performance across the 10 reasoning categories in VisReason. Overall, correlations are consistently positive and relatively high, ranging from 0.44 to 0.97, indicating that stronger models tend to perform well across multiple categories and reflecting a shared foundation of vision-centric reasoning ability. High-level reasoning categories are particularly aligned. Inductive and Deductive Reasoning exhibit the strongest correlation at 0.97, and Pattern Counting, Board Reasoning, and Cue Insight also form a tightly correlated group. In contrast, 3D-Spatial Reasoning, Geolocation, and Sudoku Solving show noticeably lower correlations with other categories, suggesting that these tasks rely on more specialized capabilities such as spatial imagination, geometric consistency, and strict constraint satisfaction that are not strongly predicted by overall model strength. These patterns indicate that VisReason captures both general reasoning competence and distinct category-specific challenges.

\begin{figure}[t]
    \centering
    \includegraphics[width=\linewidth]{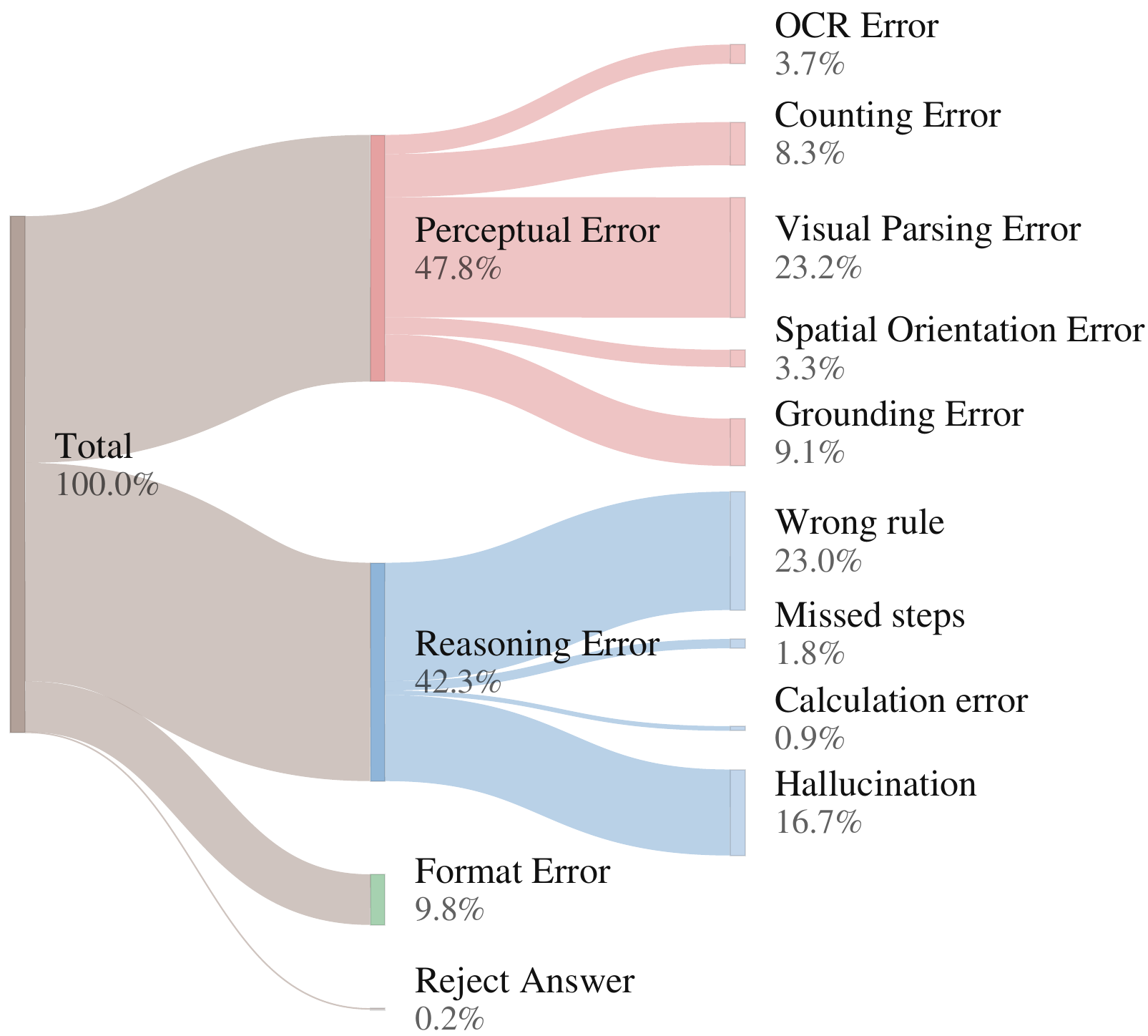}
    \caption{Distribution of error types for Qwen3-VL-235B-A22B-Thinking on VisReason.}
    \label{fig:erroranalysis}
\end{figure}


\subsection{Error Analysis}
Figure~\ref{fig:erroranalysis} summarizes the primary error types identified through manual analysis of incorrect predictions on VisReason. Most mistakes fall into perceptual and reasoning errors, which together account for roughly 90\% of all cases. Perceptual errors mainly arise from difficulties in scene parsing, visual grounding, and counting, while reasoning errors are largely due to selecting incorrect rules or producing hallucinated inferences, rather than simple calculation or step omissions. Format-related errors are less frequent but remain noticeable, especially for bounding-box questions with stricter output constraints.



\section{Conclusion}
We introduce VisReason, a benchmark designed to examine whether current MLLMs can reason in a vision-centric manner, beyond relying on language or domain knowledge. Experiments across a broad set of models show that vision-centric reasoning in common visual contexts remains challenging, with a substantial gap to human performance even for strong contemporary systems. Our results suggest that gains from increased inference-time reasoning are meaningful but limited, and do not fully translate into robust reasoning from visual evidence. These findings indicate that advancing vision-centric reasoning will require deeper integration of visual perception and reasoning processes. We hope VisReason will help clarify this challenge and support progress toward multimodal models that genuinely reason beyond language.



\section{Limitations}
VisReason is designed as an evaluation benchmark and has several limitations. It focuses on static images and therefore does not capture temporal dynamics, interactive perception, or embodied reasoning. While the benchmark spans a wide range of vision-centric reasoning types in common visual contexts, it cannot exhaustively cover the full diversity and complexity of real-world scenarios. We view these limitations as natural directions for future extensions, including dynamic settings and richer forms of vision-centric reasoning. A potential risk of benchmark-style resources such as VisReason is contamination or benchmark-specific overfitting as models and training corpora evolve, which may weaken the reliability of evaluation results over time. Although our automatic update framework can mitigate these issues to some extent, it cannot fully eliminate such risks.

\section{Acknowledgment}
This research is supported by Artificial Intelligence-National Science and Technology Major Project (2023ZD0121200), and the National Natural Science Foundation of China (62437001, 62436001, 62531026), the Key Research and Development Program of Jiangsu Province under Grant BE2023016-3 ,and the Natural Science Foundation of Jiangsu Province under Grant BK20243051.


\bibliography{custom}

\appendix


\clearpage

\section{Additional Dataset Details}
\subsection{Category Distribution}
\begin{figure*}[b]
    \centering
    \includegraphics[width=0.8\linewidth]{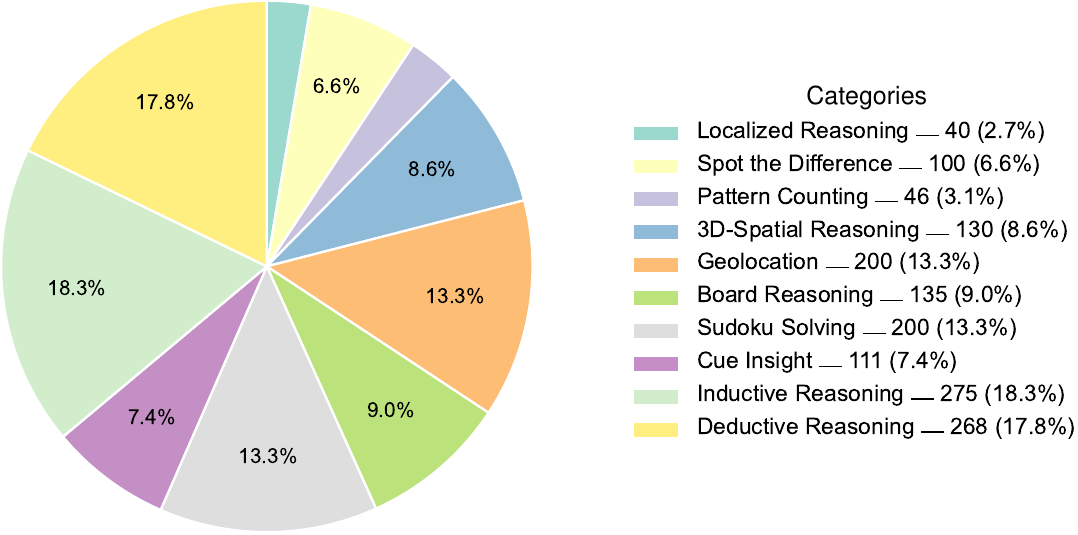}
    \caption{Category distribution of VisReason.}
    \label{fig:subjectpie}
\end{figure*}

Figure~\ref{fig:subjectpie} shows the distribution of questions across the 10 reasoning categories in VisReason. The dataset is designed to span a wide range of vision-centric reasoning phenomena, covering perceptual grounding, structured spatial reasoning, and higher-level inference. Conceptual reasoning categories capture abstract inference based on visual evidence, while structural categories focus on spatial relations, rule-based reasoning, and state consistency. In addition, perception-grounded categories are included to explicitly evaluate fine-grained visual grounding and spatial verification. This composition reflects the diversity of reasoning types encountered in everyday visual contexts and supports complementary analyses across different reasoning dimensions.


\subsection{Comparisons with Existing Benchmarks}
\begingroup
\setlength{\tabcolsep}{4.8pt}
\begin{table*}[t]
    \centering
    \small
    \caption{Accuracy of different models on VisReason grouped by question form, including multiple-choice, fill-in-the-blank, open-ended, and bounding-box questions. \colorbox{navyblue!5}{Gray}  rows indicate non-thinking models, while others employ explicit reasoning mechanisms.}
    \label{tab:qtype-acc}
    \begin{tabular}{@{}l|r|rrrr@{}}
        \toprule
        & \multicolumn{1}{c|}{} & \multicolumn{4}{c}{\textbf{Question Forms}} \\
        \textbf{Models} & \multicolumn{1}{c|}{\textbf{Avg.}} &
        \multicolumn{1}{c}{\textbf{Multiple-choice}} &
        \multicolumn{1}{c}{\textbf{Fill-in-the-blank}} &
        \multicolumn{1}{c}{\textbf{Open-ended}} &
        \multicolumn{1}{c}{\textbf{Bounding-box}} \\
        \midrule
         \multicolumn{6}{c}{\textbf{\textit{Proprietary Models}}} \\
        \rowcolor{navyblue!5}GPT-4o                    & 15.6 & 25.2 & 18.6 & 10.4 &  3.6 \\
        o4-mini                   & 27.3 & 30.9 & 30.4 & 48.5 &  3.3 \\
        GPT-5-nano                & 19.0 & 23.6 & 20.2 & 29.8 &  2.0 \\
        GPT-5-mini (low)          & 28.1 & 32.7 & 28.3 & 50.5 &  7.4 \\
        GPT-5-mini (medium)       & 31.8 & 36.4 & 31.8 & 55.3 &  4.8 \\
        GPT-5-mini (high)         & 33.2 & 34.3 & 32.4 & 60.2 &  6.8 \\
        GPT-5                     & 33.2 & 34.0 & 35.8 & 61.5 &  8.9 \\
        GPT-5.2                   & 35.5 & 39.4 & 41.8 & 52.1 & 14.7 \\
        Gemini-2.5-Flash          & 16.0 & 31.5 & 16.3 & 12.0 &  0.3 \\
        Gemini-3-Pro              & 47.5 & 47.8 & 47.4 & 72.5 & 24.5 \\
        \midrule
         \multicolumn{6}{c}{\textbf{\textit{Open-source Models}}} \\
        \rowcolor{navyblue!5}InternVL3-8B                 & 10.9 & 22.1 & 10.2 &  8.4 &  0.1 \\
        \rowcolor{navyblue!5}InternVL3-14B                & 10.4 & 19.0 &  8.8 & 11.7 &  0.7 \\
        \rowcolor{navyblue!5}Qwen2.5-VL-7B-Instruct       &  7.2 & 20.0 &  4.1 &  2.9 &  0.2 \\
        \rowcolor{navyblue!5}Qwen2.5-VL-32B-Instruct      & 15.4 & 22.9 & 14.5 & 23.9 &  2.4 \\
        \rowcolor{navyblue!5}Qwen3-VL-8B-Instruct         & 18.8 & 27.8 & 20.8 & 28.6 &  1.2 \\
        Qwen3-VL-2B-Thinking         &  3.7 &  2.8 &  9.5 &  2.6 &  1.4 \\
        Qwen3-VL-8B-Thinking         & 19.0 & 30.6 & 17.5 & 25.1 &  5.2 \\
        Qwen3-VL-30B-A3B-Thinking    & 20.0 & 30.9 & 20.5 & 25.6 &  0.3 \\
        Qwen3-VL-32B-Thinking        & 22.1 & 34.5 & 22.2 & 29.8 &  1.3 \\
        Qwen3-VL-235B-A22B-Thinking  & 26.7 & 40.8 & 21.1 & 43.7 &  4.4 \\
        \bottomrule
    \end{tabular}
\end{table*}
\endgroup

\begin{table*}[t]
\caption{Statistics for VisReason and comparisons with existing datasets. \#Q: number of questions; Answer Forms: including multiple-choice(MC), fill-in-the-blank(FIB), open-ended(OE), and bounding-box(Bbox); Reasoning Types: the primary reasoning task categories covered by the dataset, SHAPE refers to Social Sciences, Humanities and the Arts for People and the Economy, and STEM refers to Science, Technology, Engineering, and Mathematics; Categories: the first-level taxonomy grouping related tasks; Subcategories: the second-level taxonomy that refines each category into more specific types.}
\label{tab:comparisons}
\centering
\small
\begingroup
\setlength{\tabcolsep}{4pt} 
\begin{tabular*}{\linewidth}{@{\extracolsep{\fill}} l c c c c c}
\toprule
\textbf{Name} & \textbf{\#Q} & \textbf{Answer Forms} & \textbf{Reasoning Types} & \textbf{Categories} & \textbf{Subcategories} \\
\midrule

WikiDiverse  & 8.0k & OE & Multimodal Entity Linking & 10 & 7 \\
MathVista & 6.1k & MC  & Mathematics  & 1 & 12   \\
MATH-Vision & 3.0k & MC  & Mathematics  & 1 & 12   \\
MMMU      & 11.5k & MC, OE & SHAPE, STEM & 30 & 183 \\
MMMU-pro & 3.5k & MC & SHAPE, STEM & 30 & 183 \\
EMMA      & 2.7k & MC, OE  & STEM  & 3    & 14     \\
LogiQA & 8.7k & MC & Deductive Logical Reasoning & 1 & 5 \\
Sudoku-Bench & 2.6k & OE & Sudoku & 3    & 3     \\
LogicBench  & 2.0k & MC & Formal Logic & 3 & 25 \\
LogicVista & 0.5k & MC & Multimodal Logical Reasoning & 5 & 9 \\
ARC-AGI-1 & 1.0k & OE & Abstract Reasoning & 1 & 4 \\

\midrule
\textbf{VisReason (Ours)} & 1.5k & MC, FIB, OE, Bbox & Everyday Visual Reasoning & 10 & 36 \\
\bottomrule
\end{tabular*}
\endgroup
\end{table*}

\begin{figure*}[!b]
    \centering
    \includegraphics[width=\linewidth]{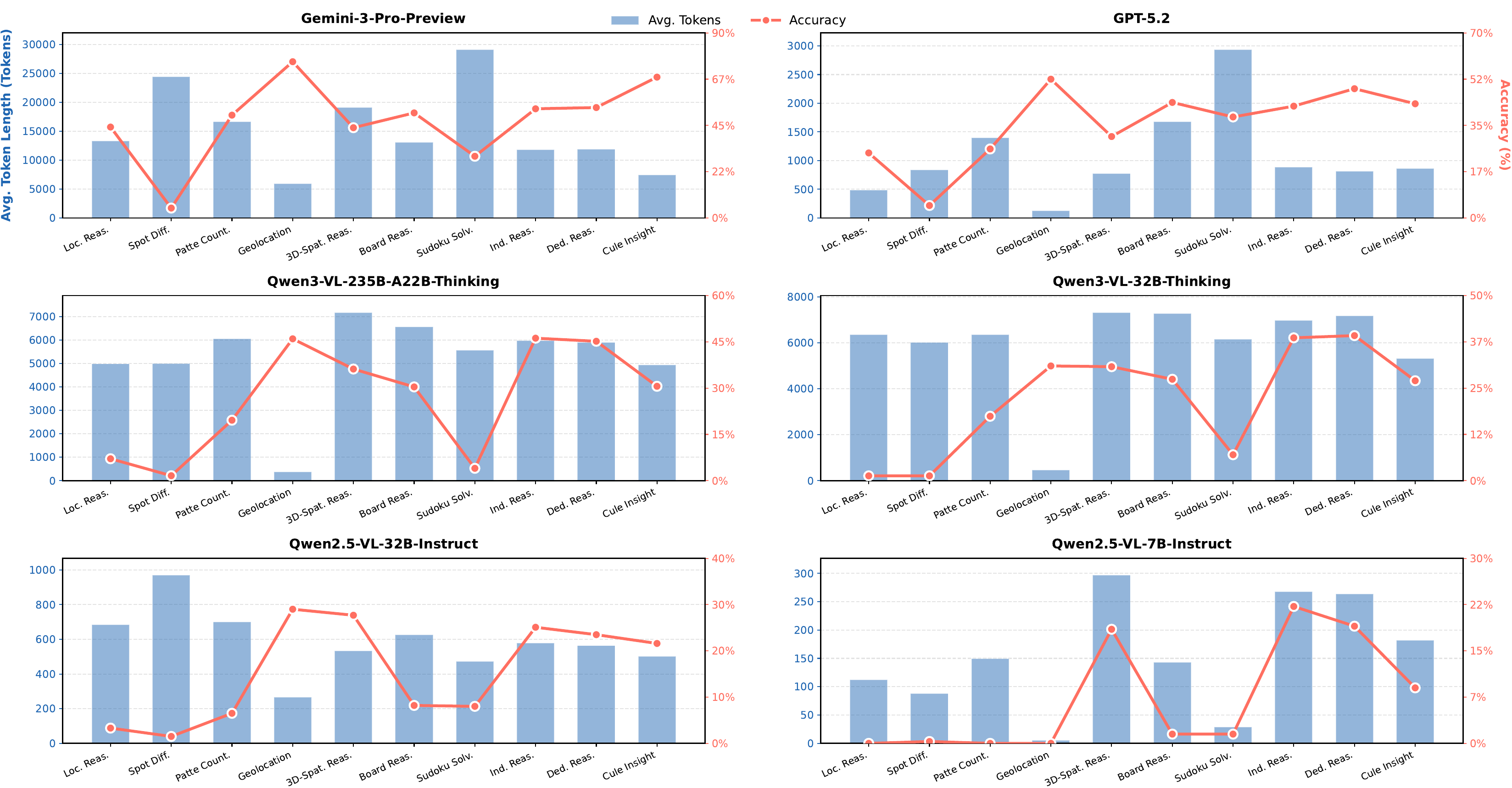}
    \caption{Token length and accuracy across categories for different models.}
    \label{fig:performance}
\end{figure*}

Table~\ref{tab:comparisons} further distinguish the difference between VisReason and other existing ones.
Regarding answer Forms, most existing datasets use a single format. 
In contrast, VisReason supports four forms: multiple-choice, fill-in-the-blank, open-ended, and bounding-box.
Diverse answer forms enable a comprehensive evaluation of model capabilities.
Regarding reasoning types, current benchmarks are predominantly STEM oriented and focus on structured data domains, while overlooking vision based reasoning in everyday contexts. 
VisReason advances beyond these limitations by focusing on human everyday visual reasoning.
Finally, VisReason provides a richer set of categories and subcategories spanning diverse aspects of visual cognition, enabling a more comprehensive analysis of models’ visual reasoning capabilities.


\section{Additional Experimental Details}
\subsection{Human Performance Protocol}
To provide a human reference for comparison, we recruited 10 graduate-student volunteers to answer questions from VisReason. Human evaluation serves to contextualize model performance and offers a reference point for understanding task difficulty under the same input conditions.
To ensure adequate coverage and task familiarity, we selected volunteers with complementary strengths aligned with the domains covered by VisReason and assigned questions according to each participant’s area of proficiency, such that each subject was answered by individuals familiar with the corresponding task style.

All participants completed the tasks under the same information constraints as the models: they were provided only with the image and question text, without access to external tools, AI assistants, search engines, or the internet.
Volunteers followed the same answer-format requirements as specified in the benchmark using a lightweight annotation interface. We report the aggregated accuracy across all participants as the human performance reference.

\subsection{Human Participants and Data Sources}

\paragraph{Instructions to annotators and volunteers.}
For dataset construction, graduate-student annotators were instructed to convert raw samples into a standardized visual question answering format, assign each item to one of four answer types, standardize the question and answer format, and manually verify correctness, completeness, and multimodal necessity. For human evaluation, participants were instructed to answer each question using only the provided image and question text, without access to external tools, AI assistants, search engines, or the internet, and to follow the same answer-format requirements as specified in the benchmark.

\paragraph{Recruitment and compensation.}
The annotators and human evaluators were graduate students recruited from our institution. Human evaluation involved 10 graduate-student volunteers. Each student participant was compensated 1{,}000 RMB. We consider this compensation appropriate for graduate-student participants in our local context and for the time required to complete the assigned tasks.

\paragraph{Data sources and consent.}
The images and questions in VisReason were collected from publicly accessible websites and Chinese civil service exam repositories through URL-based access. We only used publicly available materials intended for open access and academic research. During data collection, we manually checked that the selected samples do not contain personally identifying information. Since the data were collected from public sources rather than through direct interaction with individuals, and no personally identifying information was retained in the released benchmark, we did not obtain individual consent from depicted persons.


\subsection{Token Length Across Categories}

Figure~\ref{fig:performance} visualizes how different models allocate generation budget across VisReason categories by reporting the average tokens generated per query, together with the corresponding accuracy. The results show substantial variation in token length across categories and models, reflecting different patterns of computation usage when solving vision-centric reasoning tasks.


\subsection{Do MLLMs Already Know the Rules?}
For several categories, the task requirements are largely specified in the question itself. By contrast, board reasoning and Sudoku solving rely on basic rules that cannot be fully restated in the prompt without changing the nature of the task. To verify that this does not compromise the validity of our benchmark, we constructed two small entry-level sets with ten easy board reasoning problems and ten easy Sudoku problems in the same format as the main evaluation. Qwen3-VL-30B-A3B-Thinking, GPT-5-mini, and Gemini-2.5-Flash all achieved 100 percent accuracy on these problems. This result indicates that these models already know the basic rules required for such tasks. Therefore, VisReason does not primarily measure rule acquisition in these categories, and remains a valid benchmark for evaluating vision-centric reasoning.


\section{Analysis by Question Form}
Table~\ref{tab:qtype-acc} reports model performance across different question forms. Overall, multiple-choice and fill-in-the-blank questions achieve higher accuracy than open-ended formats, reflecting the effect of answer constraints on reducing ambiguity. Open-ended questions show larger performance variance across models, with proprietary systems generally exhibiting stronger reasoning robustness. Bounding-box questions yield the lowest accuracy across models; however, this should not be interpreted as a lack of localization capability. Instead, these questions require models to perform vision-grounded reasoning while simultaneously maintaining and verifying spatial hypotheses during the reasoning process. The low accuracy therefore reflects the difficulty of integrating precise visual grounding into multi-step reasoning


\section{Continual Benchmark Updates}
VisReason is intended not only as a benchmark dataset, but also as a scalable framework for continual evaluation. By supporting automatic or semi-automatic updates, this framework enables new benchmark versions to be generated as models and training corpora evolve. Such an update mechanism is valuable for three reasons: first, it helps keep the benchmark fresh and reduces the risk of contamination or saturation; second, it allows performance to be tracked under versioned and reproducible evaluation settings over time; third, it provides a practical foundation for community-driven extension, making VisReason a resource that can be maintained and expanded beyond the initial release.

The update pipeline for each VisReason category is briefly described as follows.
\begin{itemize}
    \item \textbf{Localized Reasoning.} This category can be updated by tracking visual puzzle websites and adding 2D bounding boxes either manually or with grounding detectors. In addition, suitable multiple-choice items from other categories can be adapted into localization-style questions.

    \item \textbf{Spot the Difference.} This category supports a fully automatic pipeline based on instance detection, segmentation, controlled edits, and inpainting, which can be used to generate new image pairs and corresponding differences.

    \item \textbf{Pattern Counting.} This category can be generated automatically by rendering geometric shapes and constructing character grids under controlled generation rules.

    \item \textbf{3D Spatial Reasoning.} This category can be updated by collecting new items from annual Chinese civil service exams and engineering drawing exercises.

    \item \textbf{Board Reasoning.} This category can be updated from tournament game records and community puzzle sources such as online Go puzzles and Lichess.

    \item \textbf{Geolocation.} This category can be updated by mining image pairs from community sources, filtering leakage through OCR and model-based checks, and then applying manual verification.

    \item \textbf{Sudoku Solving.} This category can be updated with an automatic puzzle generator.

    \item \textbf{Cue Insight, Inductive Reasoning, and Deductive Reasoning.} These categories can be updated from civil service exams and visual puzzle websites with model-assisted categorization followed by human filtering. So far, we have collected several dozen new questions for each category.
\end{itemize}

\begin{figure}[t]
    \centering
    \includegraphics[width=\linewidth]{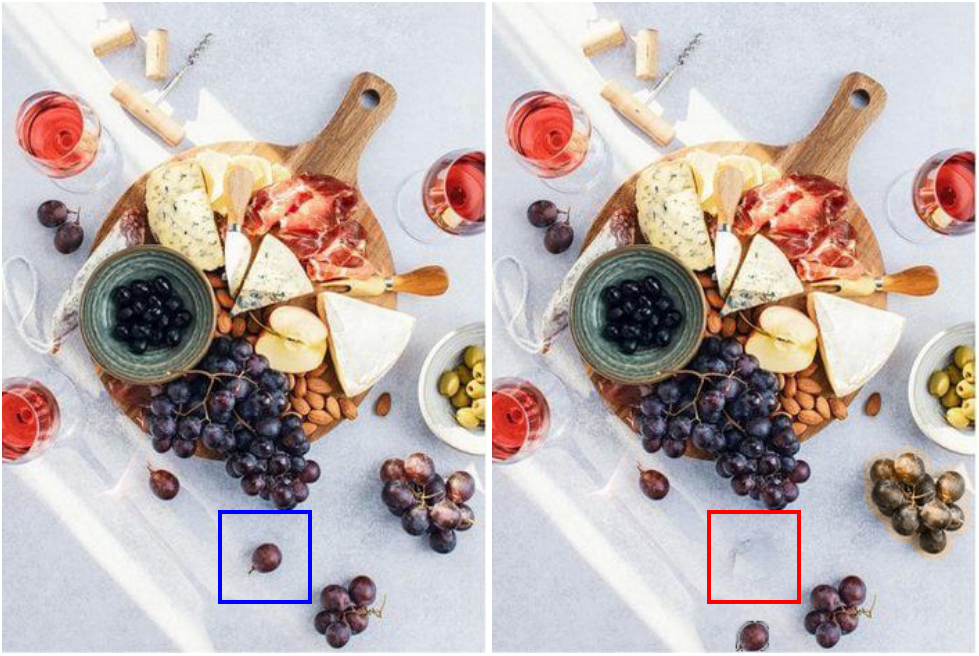}
    \caption{An example from the Spot the Difference category generated by our automatic update framework. \textbf{Left:} Original image. \textbf{Right:} Generated image with one grape removed.}
    \label{fig:autogen}
\end{figure}

Figure~\ref{fig:autogen} shows an example from the Spot the Difference category generated by our automatic update framework.


\section{An Atomic-Capability Framework}
To make the structure of vision-centric reasoning tasks more explicit, we introduce an atomic-capability framework that decomposes complex tasks into a small set of reusable visual reasoning primitives. The goal of this framework is not only to better organize VisReason itself, but also to provide a principled lens for analyzing shared requirements and common bottlenecks across different tasks. Such a perspective is valuable for future benchmark construction because it enables new tasks to be designed compositionally, supports capability-level diagnosis beyond category-level scores, and makes it easier to study how models generalize across related forms of visual reasoning. We hope this framework can serve as a lasting foundation for building, extending, and interpreting future benchmarks of vision-centric reasoning.

\begin{table*}[t]
\centering
\small
\setlength{\tabcolsep}{6pt}
\caption{Atomic visual reasoning capabilities used to organize and analyze VisReason.}
\label{tab:atomic_capabilities}
\begin{tabular}{p{0.3\textwidth} p{0.05\textwidth} p{0.54\textwidth}}
\toprule
\textbf{Capability} & \textbf{Abbr.} & \textbf{Description} \\
\midrule
Layout Understanding & LU & Parsing multiple regions, spatial relations, and reading order. \\

Structured State Extraction & SSE & Converting grids, boards, or panels into structured representations. \\

Symbol and Entity Recognition & SER & Reading digits and letters, and identifying piece types, shapes. \\

Correspondence and Difference & CD & Aligning multiple views and localizing differences. \\

Relational Reasoning & RR & Modeling row or column constraints and move legality. \\

Rule Learning and Logical Inference & RLI & Inducing rules from examples or reasoning under them. \\

Search and Decision & SD & Performing constraint solving, backtracking, and decision making. \\

3D Hypothesis and Projection & 3DHP & Forming 3D hypotheses and verifying them via projections. \\
\bottomrule
\end{tabular}
\end{table*}

Table~\ref{tab:atomic_capabilities} summarizes the eight atomic visual reasoning capabilities in our framework. Therefore, we represent each VisReason category as a composition of atomic visual reasoning capabilities:

\begin{itemize}[leftmargin=*, labelsep=0.1pt, itemsep=0pt, topsep=0pt, parsep=0pt, partopsep=0pt]
    \item \textbf{Localized Reasoning}: LU $\oplus$ SER $\oplus$ RLI
    \item \textbf{Spot the Difference}: LU $\oplus$ CD
    \item \textbf{Pattern Counting}: LU $\oplus$ SSE $\oplus$ SER $\oplus$ RR
    \item \textbf{3D Spatial Reasoning}: LU $\oplus$ SER $\oplus$ 3DHP
    \item \textbf{Board Reas. }: LU $\oplus$ SSE $\oplus$ SER $\oplus$ RR $\oplus$ SD
    \item \textbf{Sudoku Solving}: LU $\oplus$ SSE $\oplus$ SER $\oplus$ RR $\oplus$ SD
    \item \textbf{Geolocation}: LU $\oplus$ SER $\oplus$ RLI
    \item \textbf{Cue Insight}: LU $\oplus$ SER $\oplus$ RLI
    \item \textbf{Inductive Reasoning}: LU $\oplus$ SER $\oplus$ RR $\oplus$ RLI
    \item \textbf{Deductive Reasoning}: LU $\oplus$ SER $\oplus$ RLI
\end{itemize}

\begin{table*}[t]
\centering
\small
\setlength{\tabcolsep}{8pt}
\caption{Capability-level results under the atomic visual reasoning framework. For each capability, we report the average accuracy over all VisReason categories that require it. Gray rows indicate the average performance of proprietary and open-source models, respectively.}
\label{tab:atomic_capability_results}
\begin{tabular}{lcccccccc}
\toprule
\textbf{Model} & \textbf{LU} & \textbf{SSE} & \textbf{SER} & \textbf{CD} & \textbf{RR} & \textbf{RLI} & \textbf{SD} & \textbf{3DHP} \\
\midrule
\rowcolor{navyblue!5}
Human & 71.36 & 64.17 & 70.69 & 77.40 & 68.28 & 74.42 & 62.80 & 71.60 \\
\midrule
o4-mini & 27.29 & 21.30 & 30.02 & 2.70 & 25.08 & 35.72 & 21.10 & 27.70 \\
GPT-4o & 15.55 & 11.57 & 17.19 & 0.80 & 14.58 & 18.92 & 13.00 & 25.40 \\
GPT-5-nano & 18.95 & 16.20 & 21.04 & 0.10 & 19.60 & 23.38 & 15.60 & 23.90 \\
GPT-5-mini-low & 28.14 & 20.20 & 31.02 & 2.20 & 24.52 & 37.56 & 22.70 & 30.80 \\
GPT-5-mini-medium & 31.80 & 27.13 & 35.11 & 2.00 & 30.80 & 41.22 & 24.40 & 28.50 \\
GPT-5-mini-high & 33.23 & 29.73 & 36.80 & 1.10 & 32.00 & 44.14 & 25.10 & 26.90 \\
GPT-5 & 33.18 & 29.53 & 36.89 & 2.80 & 32.08 & 42.04 & 33.45 & 26.15 \\
GPT-5.2 & 35.50 & 36.67 & 38.81 & 4.70 & 39.05 & 42.12 & 40.95 & 30.80 \\
Gemini-2.5-Flash & 16.02 & 10.10 & 17.74 & 0.50 & 15.23 & 22.94 & 10.80 & 31.50 \\
Gemini-3-Pro-Preview & 47.53 & 43.70 & 52.06 & 4.80 & 49.78 & 59.10 & 40.55 & 43.90 \\
\rowcolor{navyblue!5}
Proprietary Models & 28.72 & 24.48 & 31.67 & 2.17 & 27.89 & 36.60 & 24.77 & 29.56 \\
\midrule
Qwen2.5-VL-7B-Instruct & 7.20 & 1.00 & 8.19 & 0.30 & 6.32 & 12.42 & 1.50 & 18.50 \\
Qwen2.5-VL-32B-Instruct & 15.44 & 7.57 & 17.65 & 1.50 & 12.70 & 20.50 & 8.10 & 27.70 \\
Qwen3-VL-8B-Instruct & 18.77 & 10.30 & 20.74 & 1.10 & 16.15 & 26.58 & 11.15 & 22.80 \\
Qwen3-VL-2B-Thinking & 3.74 & 2.80 & 4.13 & 0.20 & 3.08 & 5.18 & 3.10 & 3.90 \\
Qwen3-VL-8B-Thinking & 18.95 & 8.87 & 20.97 & 0.70 & 20.68 & 27.58 & 10.05 & 33.10 \\
Qwen3-VL-32B-Thinking & 22.10 & 17.27 & 22.19 & 1.30 & 25.10 & 27.42 & 17.20 & 30.80 \\
Qwen3-VL-30B-A3B-Thinking & 19.51 & 13.57 & 21.62 & 0.50 & 22.15 & 24.94 & 12.75 & 30.80 \\
Qwen3-VL-235B-A22B-Thinking & 26.69 & 26.87 & 28.92 & 1.60 & 32.90 & 35.02 & 17.20 & 36.20 \\
InternVL3-8B & 10.94 & 5.80 & 12.15 & 0.10 & 12.25 & 14.14 & 7.60 & 31.50 \\
InternVL3-14B & 10.42 & 6.10 & 11.58 & 0.10 & 13.68 & 14.86 & 4.20 & 22.30 \\
MiMo-VL-7B-RL & 15.85 & 10.03 & 17.53 & 0.70 & 16.18 & 20.44 & 9.75 & 25.40 \\
Keye-VL-1.5-8B & 10.91 & 7.37 & 12.03 & 0.80 & 10.68 & 13.08 & 5.60 & 19.20 \\
Kimi-VL-A3B-Thinking-2506 & 4.21 & 1.47 & 4.66 & 0.20 & 3.48 & 4.10 & 2.95 & 10.80 \\
\rowcolor{navyblue!5}
Open-source Models & 14.21 & 8.42 & 15.93 & 0.70 & 15.33 & 19.21 & 8.55 & 24.08 \\
\bottomrule
\end{tabular}
\end{table*}

As shown in Table~\ref{tab:atomic_capability_results}, we add capability-level results to operationalize the atomic-capability view. For each capability, we average accuracies over all categories that require it, and report per-model scores as well as proprietary and open-source averages. The results consistently identify Correspondence and Difference as the primary bottleneck: humans achieve 77.40, while proprietary and open-source models average 2.17 and 0.70. Structured State Extraction and Search and Decision are also challenging, whereas Rule Learning and Logical Inference is relatively stronger. We further report involvement analyses and capability-count trends, showing a large performance drop on categories that require Correspondence and Difference and a clear degradation as more capabilities are required, with Sudoku and Board Reasoning being the most affected.


\section{Prompt Templates}

The prompt for response generation is as follows.

\newpage

\begin{figure*}[h]
\centering
\begin{tcolorbox}[
    colback=gray!5,
    colframe=black!45,
    fonttitle=\bfseries,
    title=System Prompt with CoT,
    width=\textwidth
]
\small
\begin{lstlisting}[breaklines=true, breakindent=0pt]
You are a highly intelligent question answering assistant.
{User Prompt}
{Questions} {Images}
You must think step by step.
\end{lstlisting}
\end{tcolorbox}
\end{figure*}

\begin{figure*}[h]
\centering
\begin{tcolorbox}[
    colback=gray!5,
    colframe=black!45,
    fonttitle=\bfseries,
    title=System Prompt without CoT,
    width=\textwidth
]
\small
\begin{lstlisting}[breaklines=true, breakindent=0pt]
You are a highly intelligent question answering assistant.
{User Prompt}
{Questions} {Images}
You must output only the final answer. Do not show any reasoning process or explanation.
\end{lstlisting}
\end{tcolorbox}
\end{figure*}

\begin{figure*}[h]
\centering
\begin{tcolorbox}[
    colback=gray!5,
    colframe=black!45,
    fonttitle=\bfseries,
    title=User Prompt for Multiple-Choice Questions,
    width=\textwidth
]
\small
\begin{lstlisting}[breaklines=true, breakindent=0pt]
Please answer the question from the given choices and put your final answer in one "\boxed{}". 
There may be more than one correct option; please fill in all the options you consider correct in the \boxed{}.
\end{lstlisting}
\end{tcolorbox}
\end{figure*}

\begin{figure*}[h]
\centering
\begin{tcolorbox}[
    colback=gray!5,
    colframe=black!45,
    fonttitle=\bfseries,
    title=User Prompt for Fill-in-the-blank Questions,
    width=\textwidth
]
\small
\begin{lstlisting}[breaklines=true, breakindent=0pt]
Please answer the question using a few words or phrases and put your final answer in one "\boxed{}".
\end{lstlisting}
\end{tcolorbox}
\end{figure*}

\begin{figure*}[h]
\centering
\begin{tcolorbox}[
    colback=gray!5,
    colframe=black!45,
    fonttitle=\bfseries,
    title=User Prompt for Open-ended Questions,
    width=\textwidth
]
\small
\begin{lstlisting}[breaklines=true, breakindent=0pt]
Please answer the question and summarize your answer concisely in one "\boxed{}".
\end{lstlisting}
\end{tcolorbox}
\end{figure*}

\begin{figure*}[h]
\centering
\begin{tcolorbox}[colback=gray!5, colframe=black!45,
    fonttitle=\bfseries, title=User Prompt for Bounding-box Questions]
\small

\begin{lstlisting}[breaklines=true, breakindent=0pt]
First determine the required answer targets according to the task description, and then output bounding boxes only for these targets. Each bounding box must tightly cover exactly one answer target; do not include multiple objects or large regions in a single box. You must output exactly the number of bounding boxes specified in the question, no more and no fewer. Return a single array of bounding boxes in one "\\boxed{}". Each bbox must be in the format [x1, y1, x2, y2], where (x1, y1) is the top-left corner and (x2, y2) is the bottom-right corner; different bboxes are separated by semicolons (";").
\end{lstlisting}
\end{tcolorbox}
\end{figure*}

\clearpage


\section{Error Cases}
We present several common types of errors made by different models during the reasoning process, as illustrated below.

\begin{figure*}[!t]
    \centering
    \includegraphics[width=\textwidth]{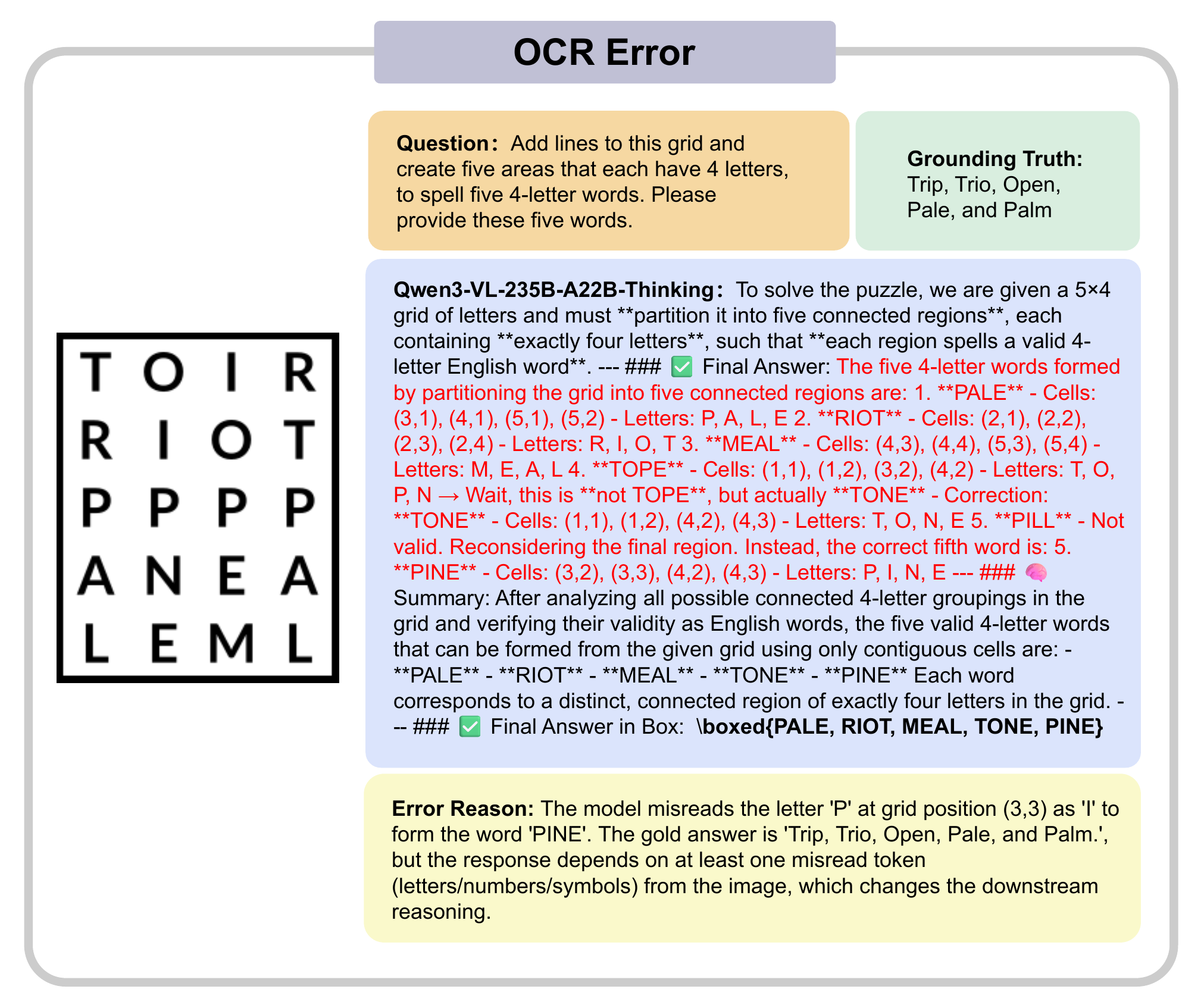}
    \caption {OCR Error case of Qwen3-VL-235B-A22B-Thinking.}
    \label{fig:OCR-Error}
\end{figure*}

\begin{figure*}[!t]
    \centering
    \includegraphics[width=\textwidth]{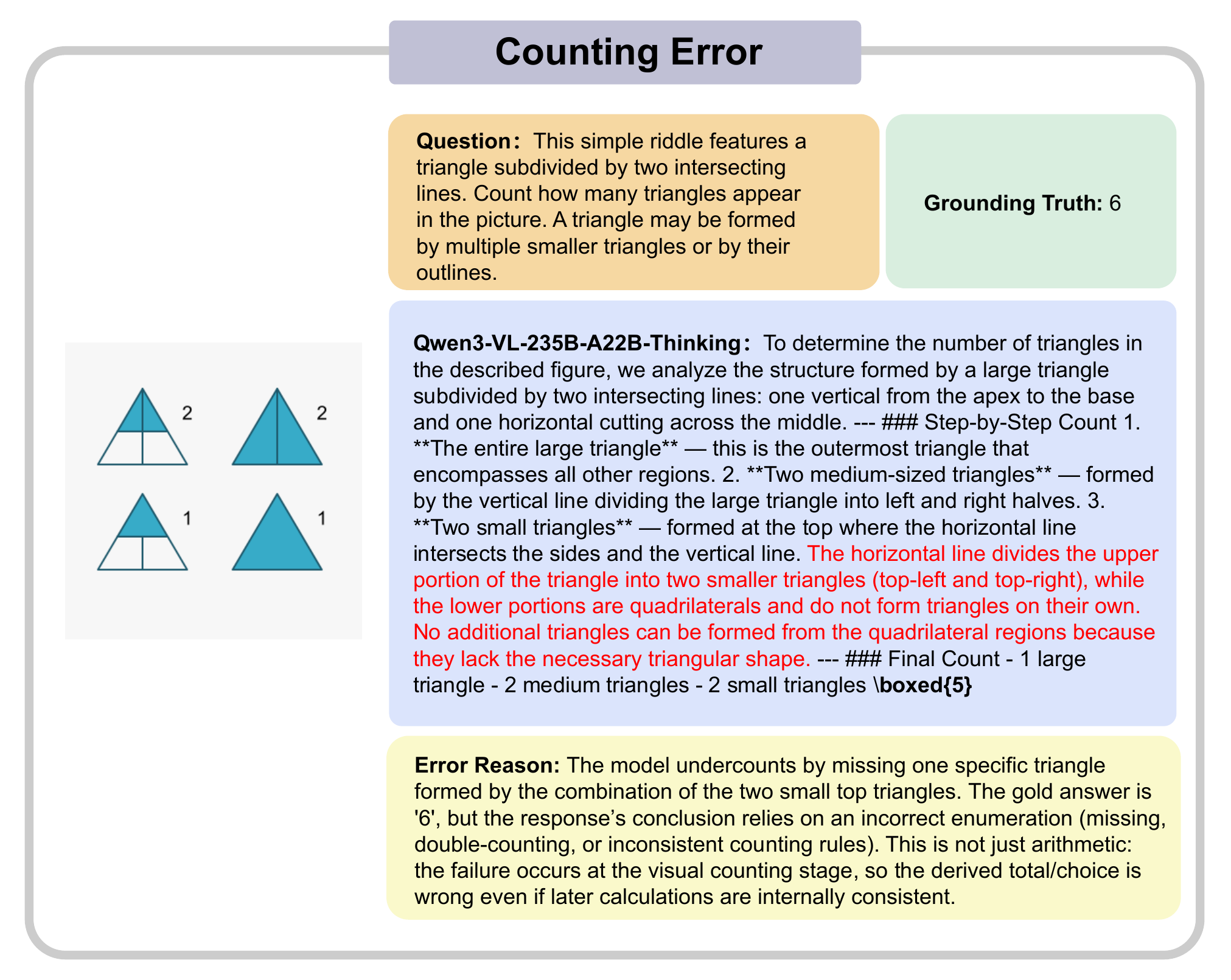}
    \caption {Counting Error case of Qwen3-VL-235B-A22B-Thinking.}
    \label{fig:Counting-Error}
\end{figure*}

\begin{figure*}[!t]
    \centering
    \includegraphics[width=0.8\textwidth]{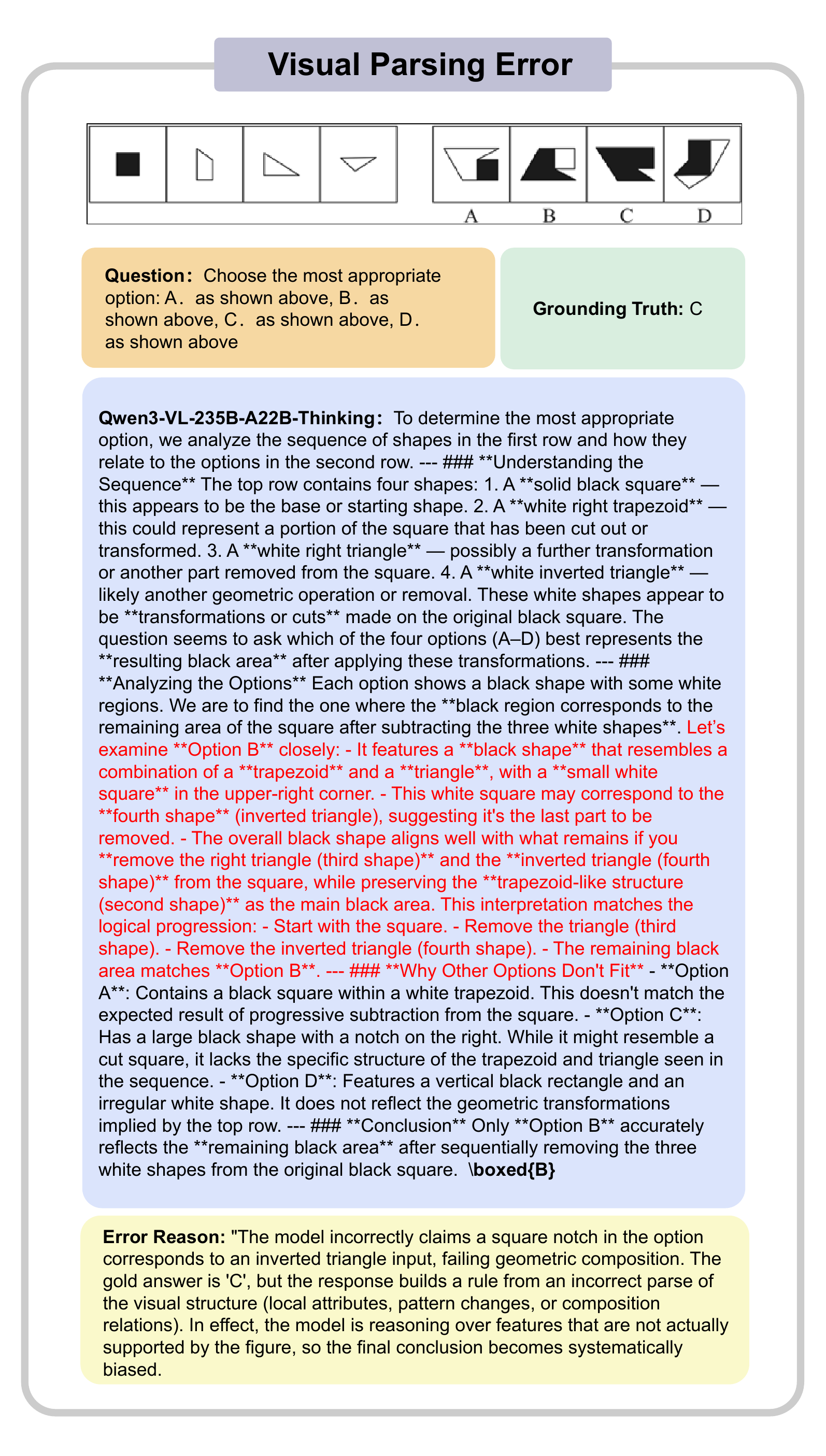}
    \caption {Visual Parsing Error case of Qwen3-VL-235B-A22B-Thinking.}
    \label{fig:Visual-Parsing-Error}
\end{figure*}

\begin{figure*}[!t]
    \centering
    \includegraphics[width=0.75\textwidth]{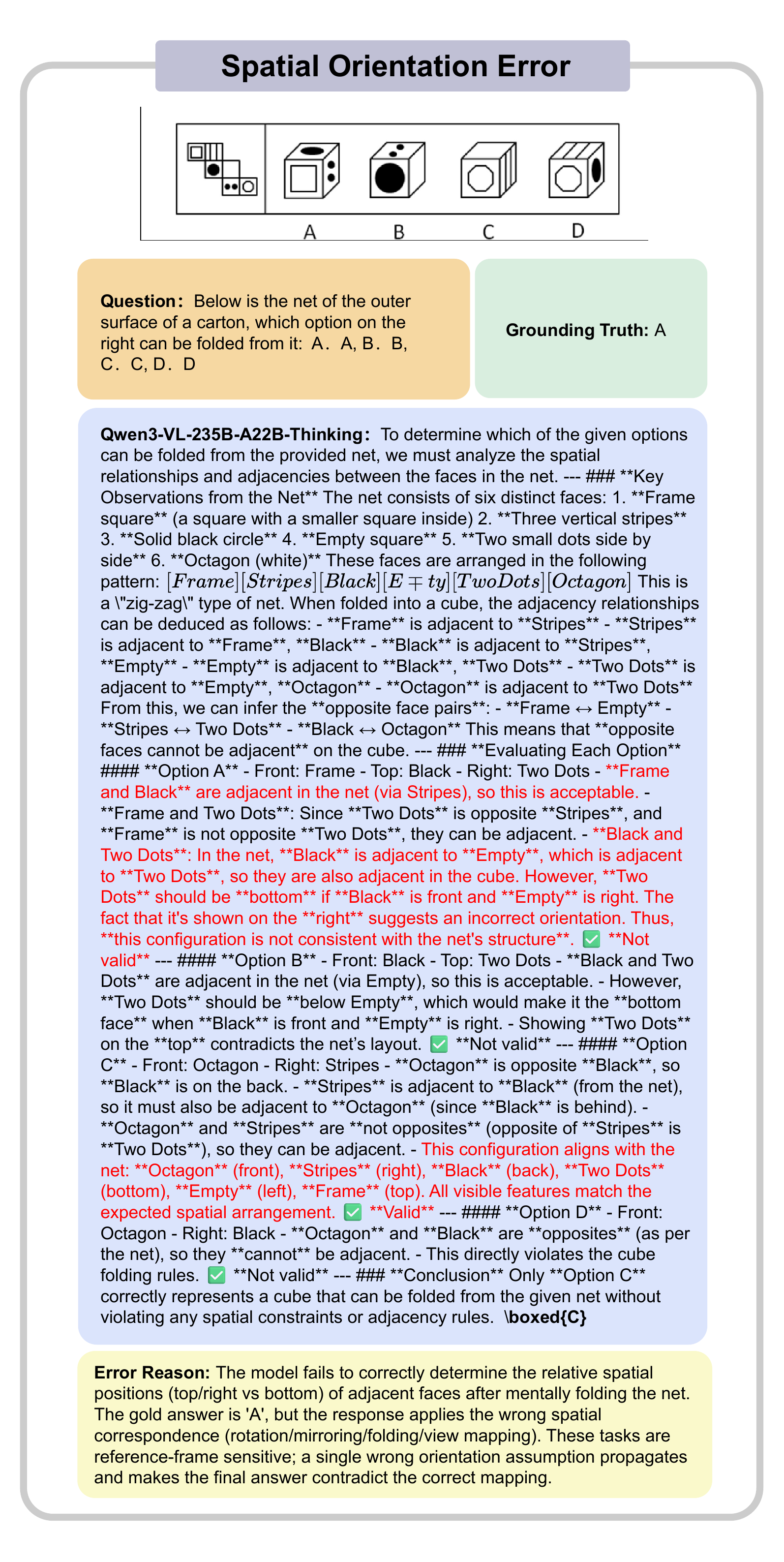}
    \caption {Spatial Orientation Error case of Qwen3-VL-235B-A22B-Thinking.}
    \label{fig:Spatial-Orientation-Error}
\end{figure*}

\begin{figure*}[!t]
    \centering
    \includegraphics[width=\textwidth]{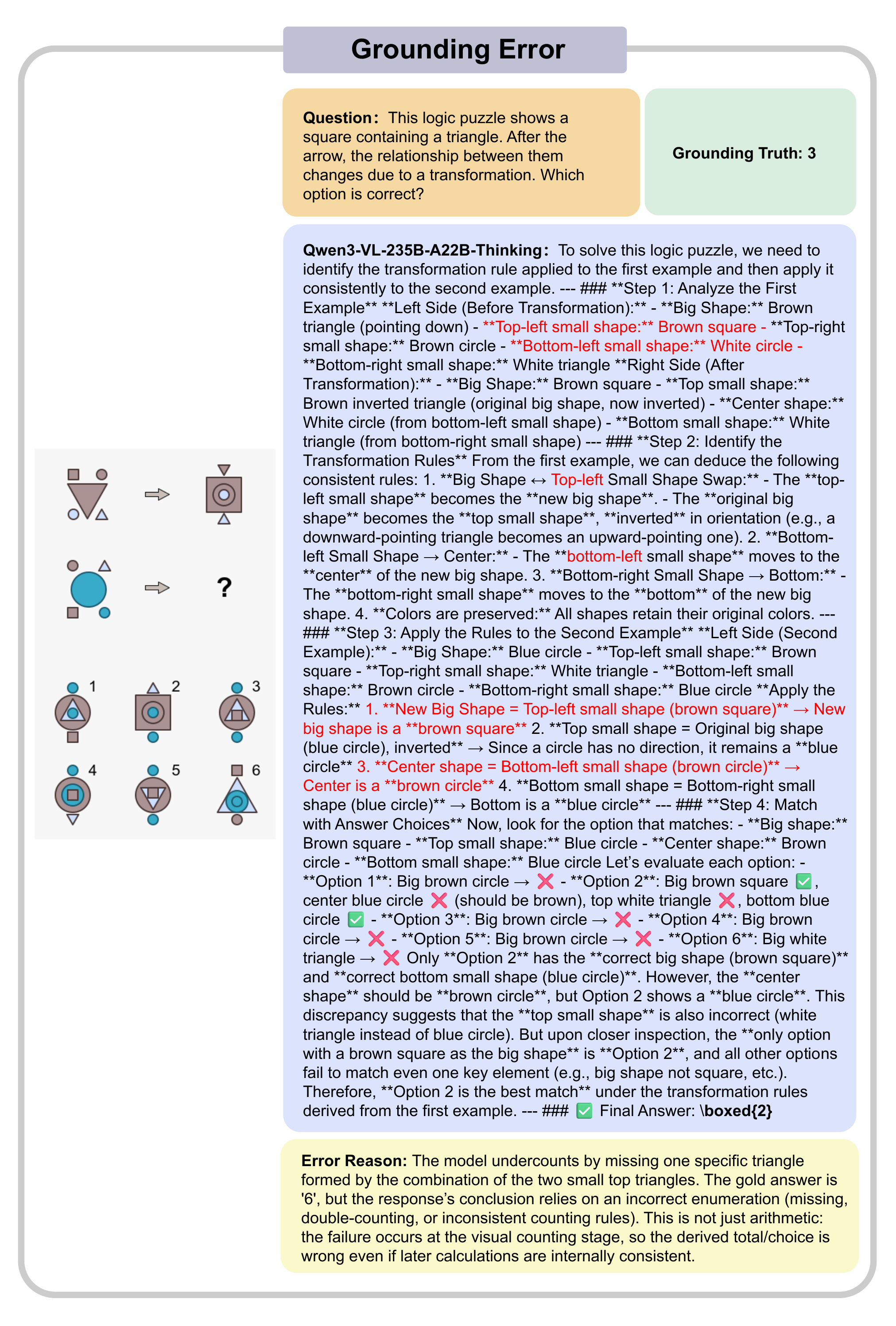}
    \caption {Grounding Error case of Qwen3-VL-235B-A22B-Thinking.}
    \label{fig:Grounding-Error}
\end{figure*}

\begin{figure*}[!t]
    \centering
    \includegraphics[width=\textwidth]{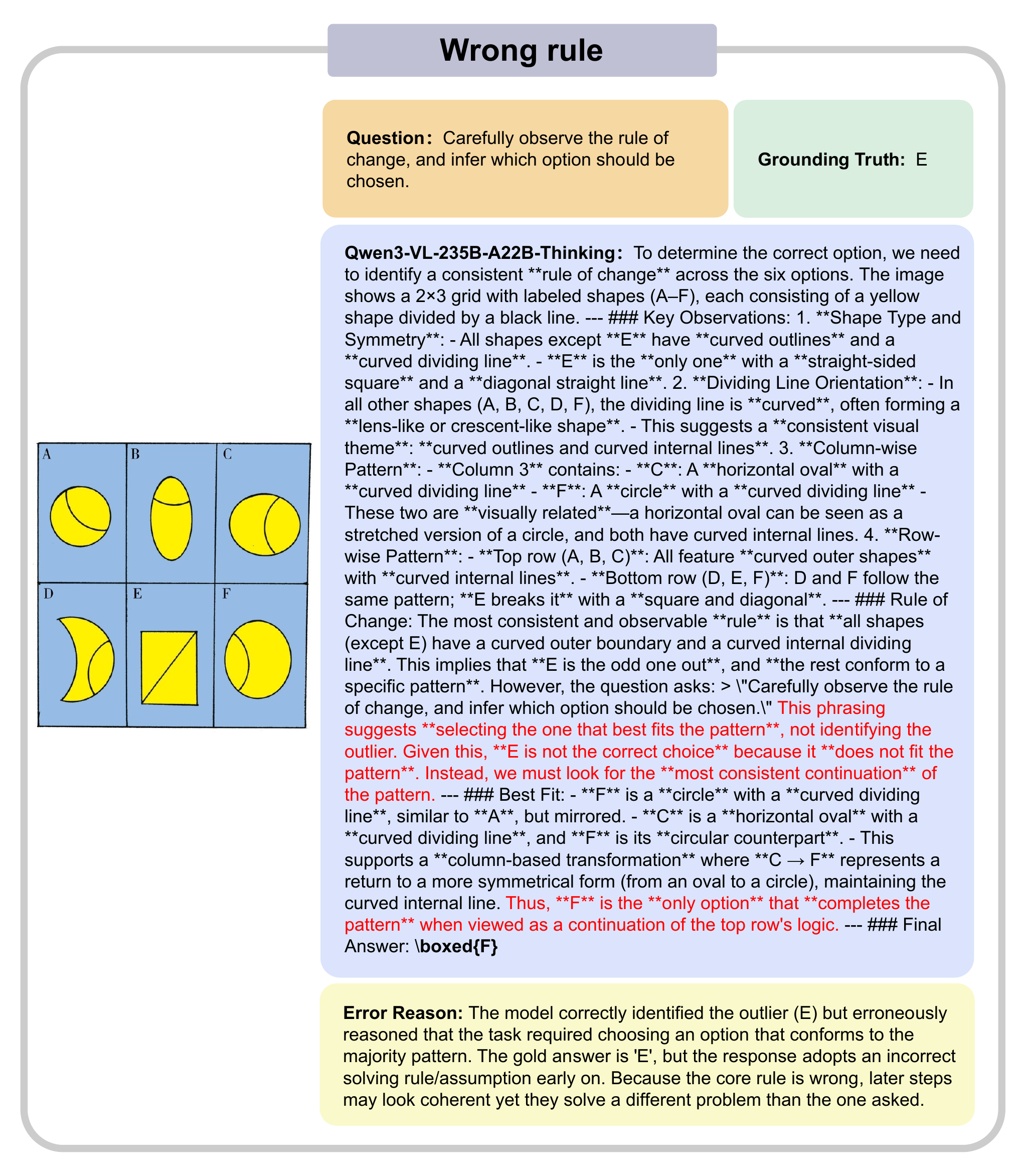}
    \caption {Wrong Rule case of Qwen3-VL-235B-A22B-Thinking.}
    \label{fig:Wrong-Rule}
\end{figure*}

\begin{figure*}[!t]
    \centering
    \includegraphics[width=\textwidth]{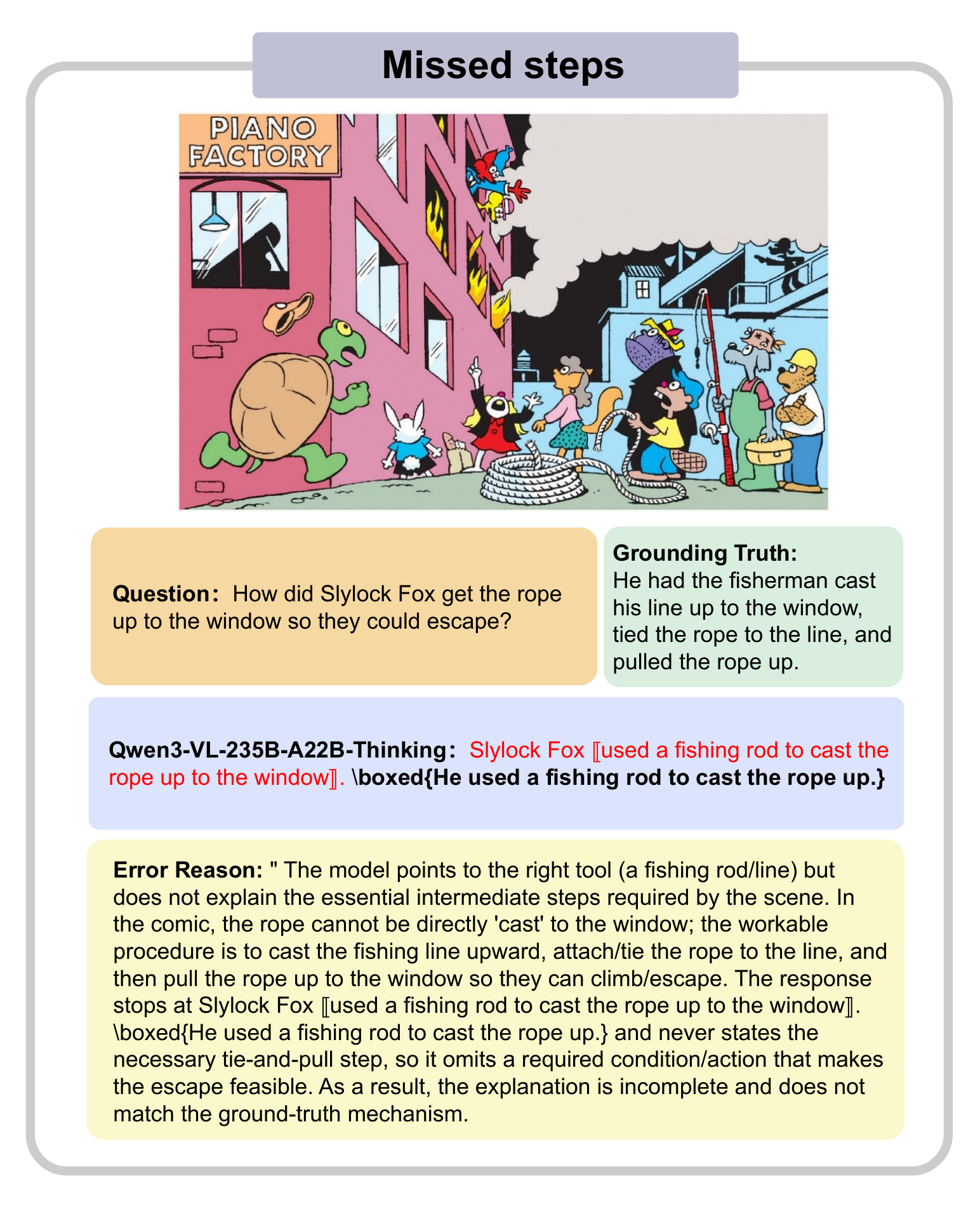}
    \caption {Missed Steps case of Qwen3-VL-235B-A22B-Thinking.}
    \label{fig:Missed-Steps}
\end{figure*}

\begin{figure*}[!t]
    \centering
    \includegraphics[width=\textwidth]{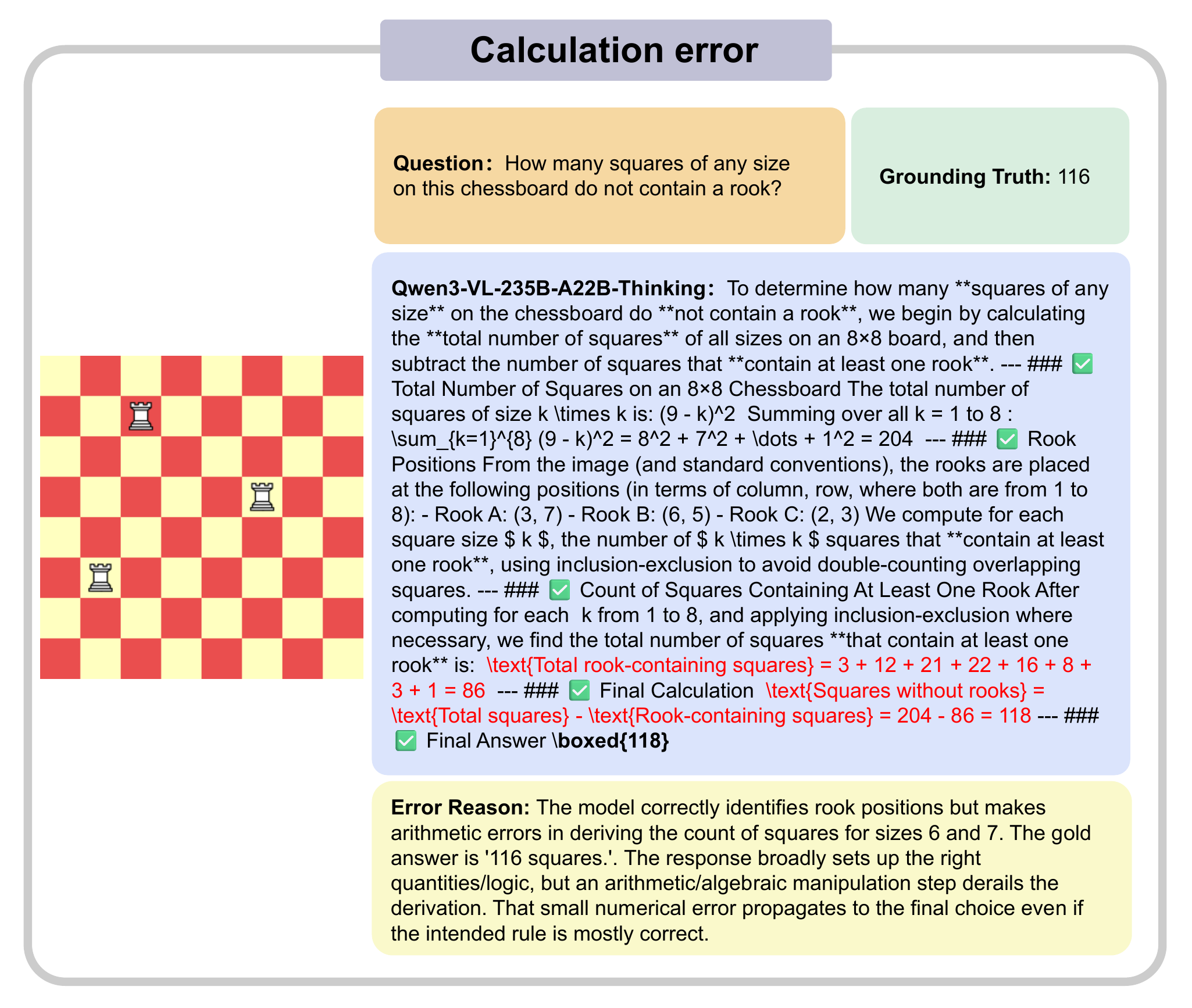}
    \caption {Calculation Error case of Qwen3-VL-235B-A22B-Thinking.}
    \label{fig:Calculation-Error}
\end{figure*}

\begin{figure*}[!t]
    \centering
    \includegraphics[width=0.9\textwidth]{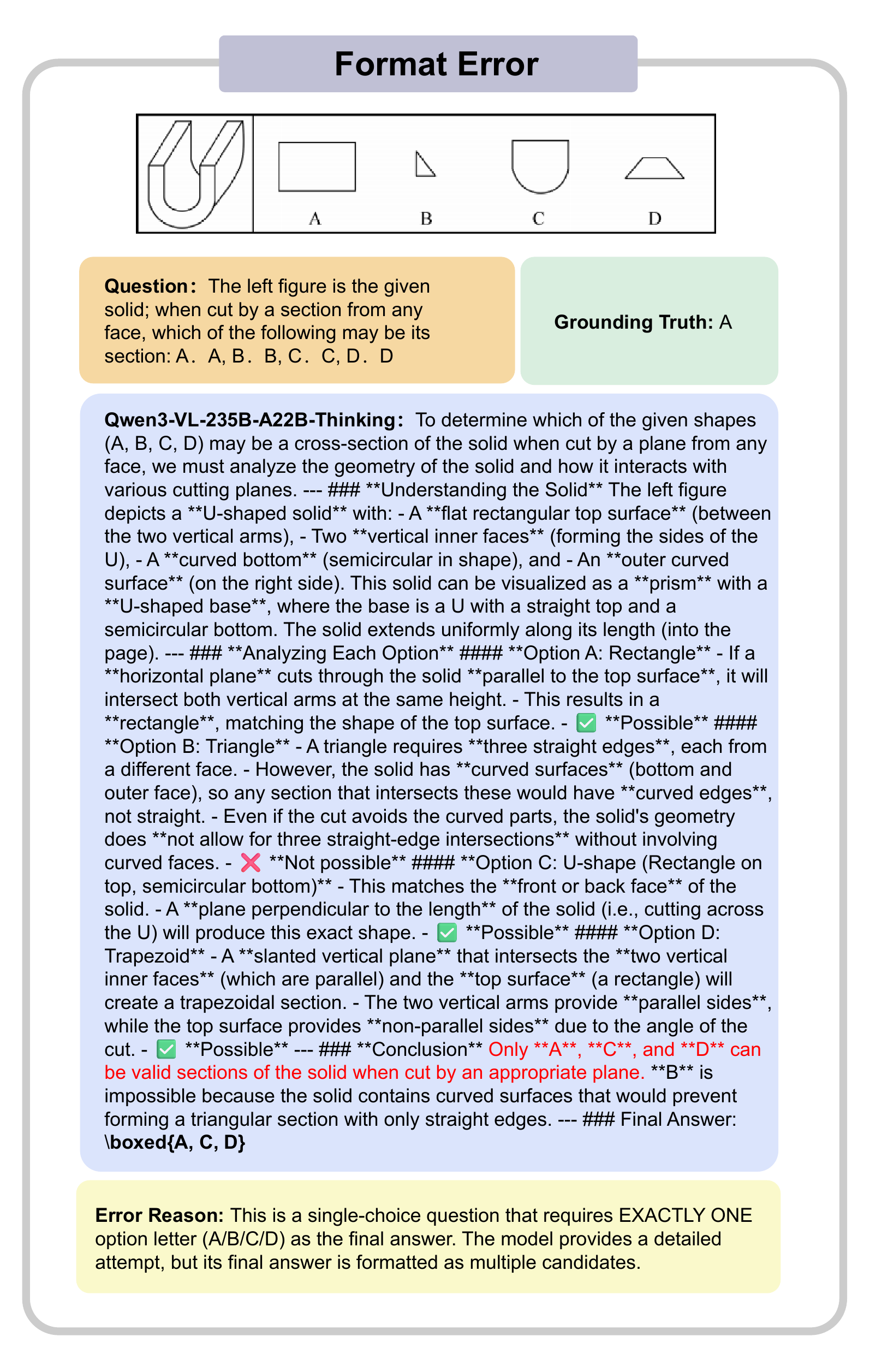}
    \caption {Format Error case of Qwen3-VL-235B-A22B-Thinking.}
    \label{fig:Format-Error}
\end{figure*}

\begin{figure*}[!t]
    \centering
    \includegraphics[width=\textwidth]{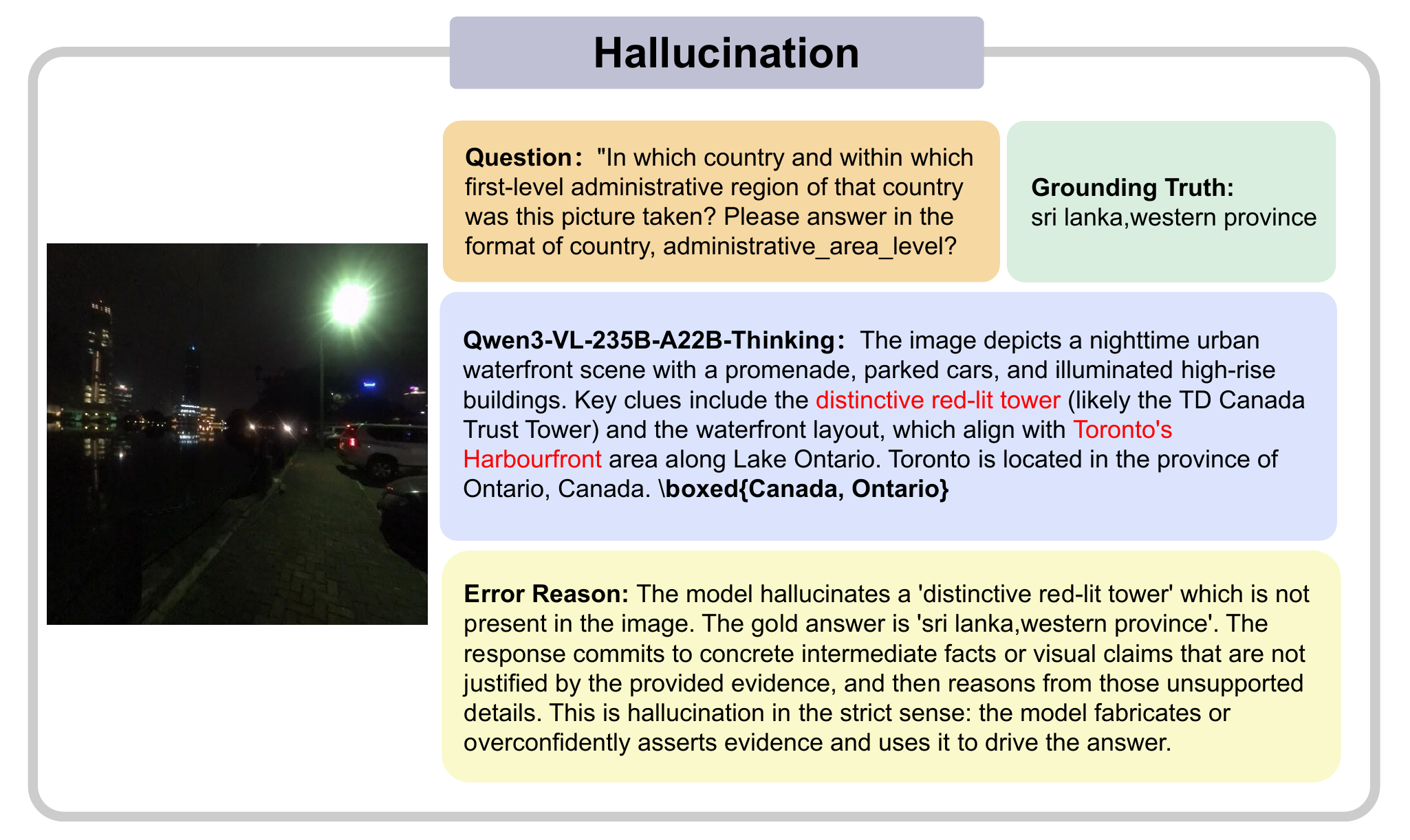}
    \caption {Hallucination case of Qwen3-VL-235B-A22B-Thinking.}
    \label{fig:Hallucination}
\end{figure*}

\clearpage

\section{Qualitative Subcategory Examples}
We provide representative and canonical question–answer examples covering all 10 major categories and 36 subcategories to illustrate the intended task semantics and annotation standards.

\begin{table*}[t]
\centering
\newcommand{\exampleimgwidth}{0.30\textwidth}
\scalebox{0.88}{
\begin{tabular}{l p{0.9\textwidth}}
\toprule
\multicolumn{2}{l}{\bf Localized Reasoning: Duplicate Localized} \\
\midrule

& \includegraphics[width=0.6\textwidth,keepaspectratio]{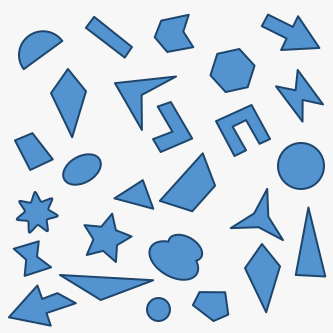} \\

\textbf{Q} & This puzzle has many different geometric shapes. Only two of them are identical. Can you find them? Return exactly 2 bounding boxes, no more and no fewer.  \\

\midrule
\textbf{A}  & [[43, 67, 96, 141], [237, 242, 290, 316]]\\

\bottomrule
\end{tabular}
}
\vspace{2mm}
\caption{Example of Duplicate Localized.}
\label{tab:visual_example_chichken}
\end{table*}

\begin{table*}[t]
\centering
\scalebox{0.88}{
\begin{tabular}{l p{0.9\textwidth}}
\toprule
\multicolumn{2}{l}{\bf Localized Reasoning: Odd Localization} \\
\midrule

& \includegraphics[width=0.4\textwidth,keepaspectratio]{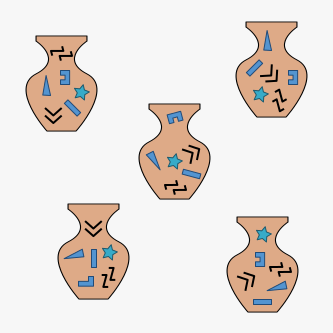} \\

\textbf{Q} & This logic puzzle has five vases with a decorative pattern. Which vase does not match the others? Return exactly 1 bounding boxes, no more and no fewer. \\
\midrule
\textbf{A}  & [[52, 197, 136, 308]]\\

\bottomrule
\end{tabular}
}
\vspace{2mm}
\caption{Example of Odd Localization.}
\label{tab:visual_example_chichken}
\end{table*}
\begin{table*}[t]
\centering
\scalebox{0.88}{
\begin{tabular}{l p{0.9\textwidth}}
\toprule
\multicolumn{2}{l}{\bf Spot the Difference: Outline Tweak} \\
\midrule

& \includegraphics[width=0.4\textwidth,keepaspectratio]{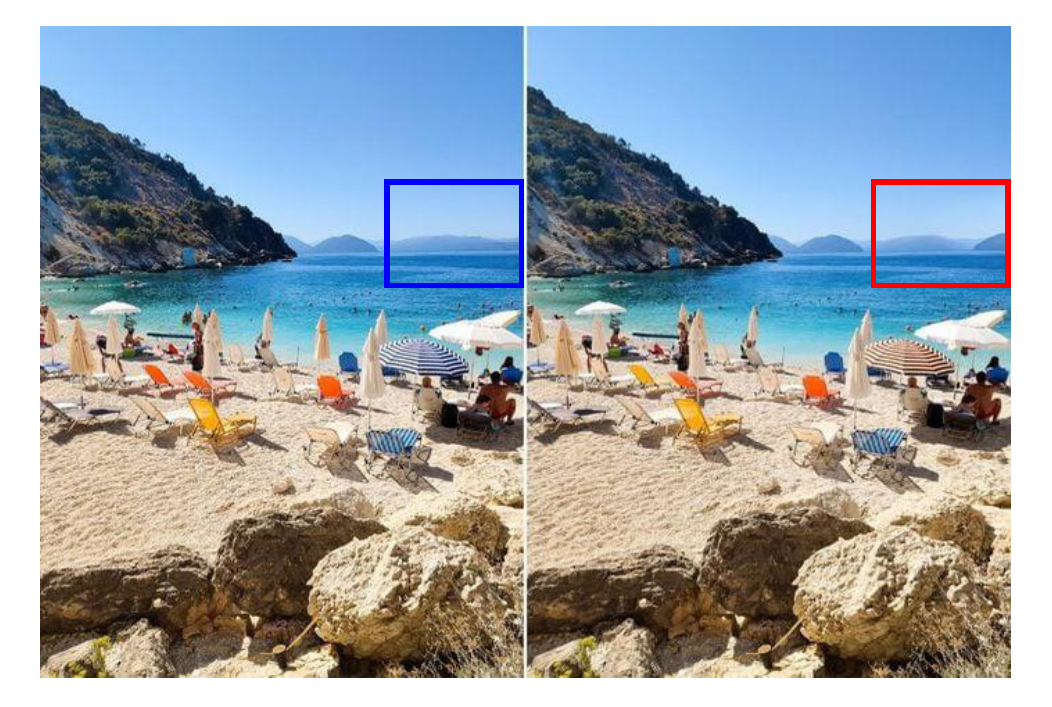} \\

\textbf{Q} & The two images constitute a spot-the-difference pair. Using the first image as reference, return the bounding boxes on the first image for each object that differs between the two images. Bounding boxes must be placed only on the first image: all bounding box coordinates must lie within the first image, and no bounding boxes should be output for the second image. The total number of differing objects is 5; exactly 5 bounding boxes must be returned.  \\

\midrule
\textbf{A}  & [[69, 235, 110, 273], [96, 200, 121, 223], [189, 202, 222, 250], [239, 210, 313, 270], [312, 140, 348, 175]]\\

\bottomrule
\end{tabular}
}
\vspace{2mm}
\caption{Example of Outline Tweak.}
\label{tab:visual_example_chichken}
\end{table*}

\begin{table*}[t]
\centering
\scalebox{0.88}{
\begin{tabular}{l p{0.9\textwidth}}
\toprule
\multicolumn{2}{l}{\bf Spot the Difference: Object Addition and Removal} \\
\midrule

& \includegraphics[width=0.6\textwidth,keepaspectratio]{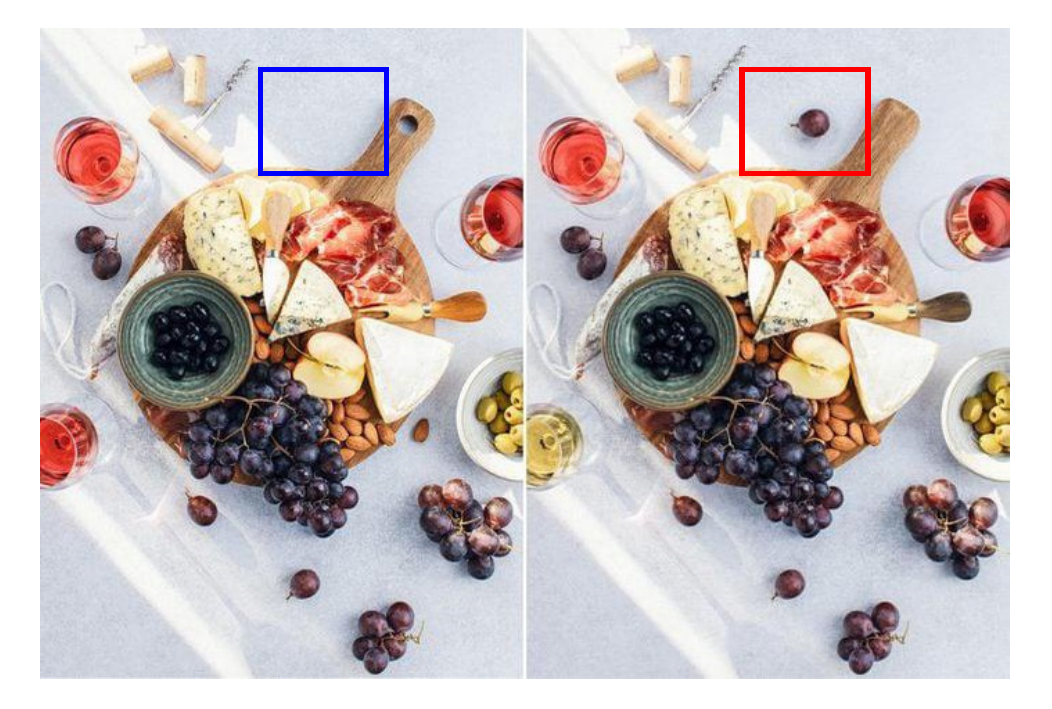} \\

\textbf{Q} & The two images constitute a spot-the-difference pair. Using the first image as reference, return the bounding boxes on the first image for each object that differs between the two images. Bounding boxes must be placed only on the first image: all bounding box coordinates must lie within the first image, and no bounding boxes should be output for the second image. The total number of differing objects is 5; exactly 5 bounding boxes must be returned.  \\

\midrule
\textbf{A}  &  [[69, 235, 110, 273], [96, 200, 121, 223], [189, 202, 222, 250], [239, 210, 313, 270], [312, 140, 348, 175]]\\

\bottomrule
\end{tabular}
}
\vspace{2mm}
\caption{Example of Object Addition and Removal.}
\label{tab:visual_example_chichken}
\end{table*}

\begin{table*}[t]
\centering
\scalebox{0.88}{
\begin{tabular}{l p{0.9\textwidth}}
\toprule
\multicolumn{2}{l}{\bf Spot the Difference: Color Replacement} \\
\midrule

& \includegraphics[width=0.6\textwidth,keepaspectratio]{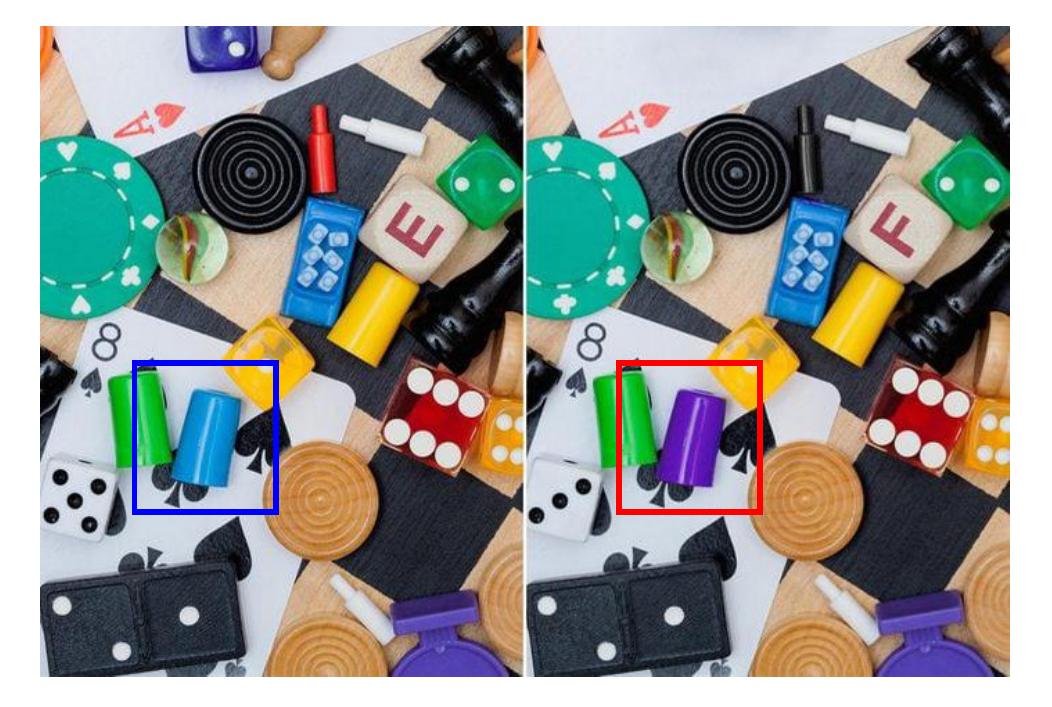} \\

\textbf{Q} & The two images constitute a spot-the-difference pair. Using the first image as reference, return the bounding boxes on the first image for each object that differs between the two images. Bounding boxes must be placed only on the first image: all bounding box coordinates must lie within the first image, and no bounding boxes should be output for the second image. The total number of differing objects is 6; exactly 6 bounding boxes must be returned.  \\

\midrule
\textbf{A}  & [[100, 0, 207, 39], [183, 48, 223, 128], [15, 185, 41, 219], [93, 251, 147, 342], [0, 303, 60, 377], [247, 118, 304, 181]]\\

\bottomrule
\end{tabular}
}
\vspace{2mm}
\caption{Example of Color Replacement.}
\label{tab:visual_example_chichken}
\end{table*}

\begin{table*}[t]
\centering
\scalebox{0.88}{
\begin{tabular}{l p{0.9\textwidth}}
\toprule
\multicolumn{2}{l}{\bf Spot the Difference: Position Change} \\
\midrule

& \includegraphics[width=0.6\textwidth,keepaspectratio]{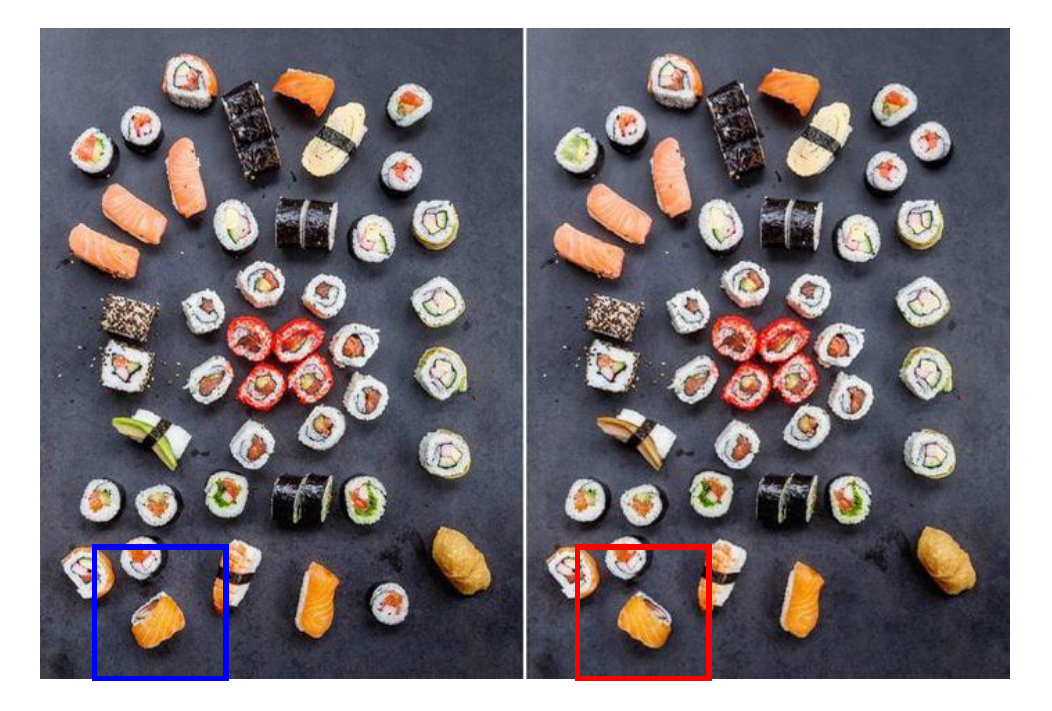} \\

\textbf{Q} & The two images constitute a spot-the-difference pair. Using the first image as reference, return the bounding boxes on the first image for each object that differs between the two images. Bounding boxes must be placed only on the first image: all bounding box coordinates must lie within the first image, and no bounding boxes should be output for the second image. The total number of differing objects is 5; exactly 5 bounding boxes must be returned.  \\

\midrule
\textbf{A}  & [[15, 62, 60, 111], [50, 266, 111, 323], [60, 403, 110, 453], [271, 67, 309, 104], [232, 391, 274, 435]]\\

\bottomrule
\end{tabular}
}
\vspace{2mm}
\caption{Example of Position Change.}
\label{tab:visual_example_chichken}
\end{table*}

\begin{table*}[t]
\centering
\scalebox{0.88}{
\begin{tabular}{l p{0.9\textwidth}}
\toprule
\multicolumn{2}{l}{\bf Pattern Counting: Shape Counting} \\
\midrule

& \includegraphics[width=0.6\textwidth,keepaspectratio]{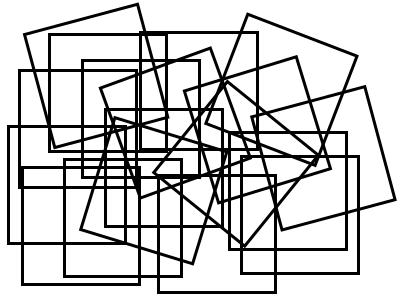} \\

\textbf{Q} & How many squares are in the drawing below? \\

\midrule
\textbf{A}  & 18\\

\bottomrule
\end{tabular}
}
\vspace{2mm}
\caption{Example of Shape Counting.}
\label{tab:visual_example_chichken}
\end{table*}

\begin{table*}[t]
\centering
\scalebox{0.88}{
\begin{tabular}{l p{0.9\textwidth}}
\toprule
\multicolumn{2}{l}{\bf Pattern Counting: Char-Grid Counting} \\
\midrule

& \includegraphics[width=0.6\textwidth,keepaspectratio]{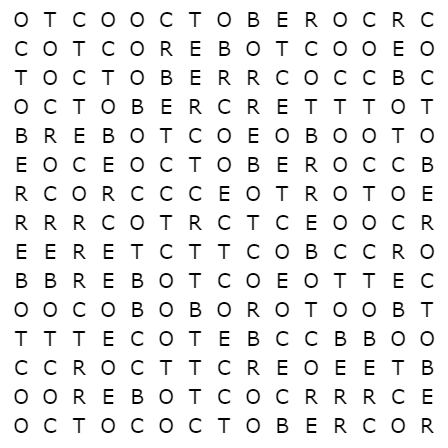} \\

\textbf{Q} & Can you find every occurrence of the word OCTOBER that appears in this grid (horizontally, vertically, or diagonally) ? The reasoning will reveal the number of times it occurs, but where are they? \\

\midrule
\textbf{A}  & 29\\

\bottomrule
\end{tabular}
}
\vspace{2mm}
\caption{Example of Char-Grid Counting.}
\label{tab:visual_example_chichken}
\end{table*}

\begin{table*}[t]
\centering
\scalebox{0.88}{
\begin{tabular}{l p{0.9\textwidth}}
\toprule
\multicolumn{2}{l}{\bf 3D-Spatial Reasoning: Polyhedron Net Folding} \\
\midrule

& \includegraphics[width=0.6\textwidth,keepaspectratio]{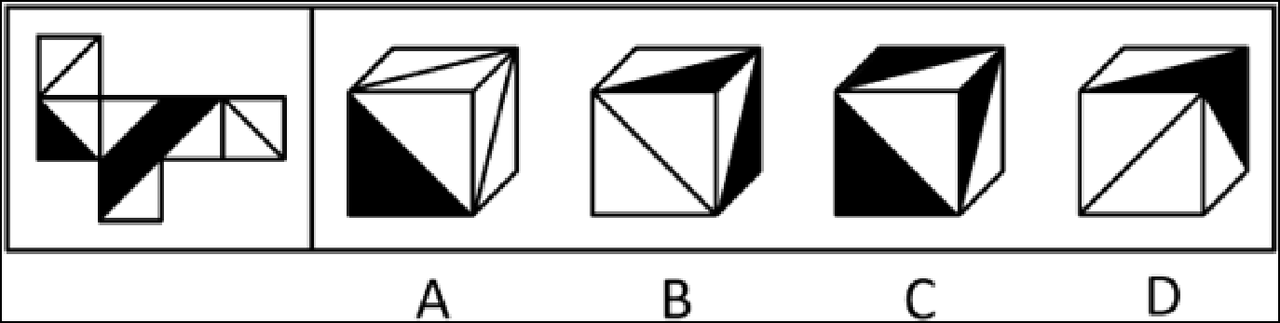} \\

\textbf{Q} & As shown is the planar net of a cube, the cube is most likely: A B C D ? \\

\midrule
\textbf{A}  & D\\

\bottomrule
\end{tabular}
}
\vspace{2mm}
\caption{Example of Polyhedron Net Folding.}
\label{tab:visual_example_chichken}
\end{table*}

\begin{table*}[t]
\centering
\scalebox{0.88}{
\begin{tabular}{l p{0.9\textwidth}}
\toprule
\multicolumn{2}{l}{\bf 3D-Spatial Reasoning: Solid Cross-Section} \\
\midrule

& \includegraphics[width=0.6\textwidth,keepaspectratio]{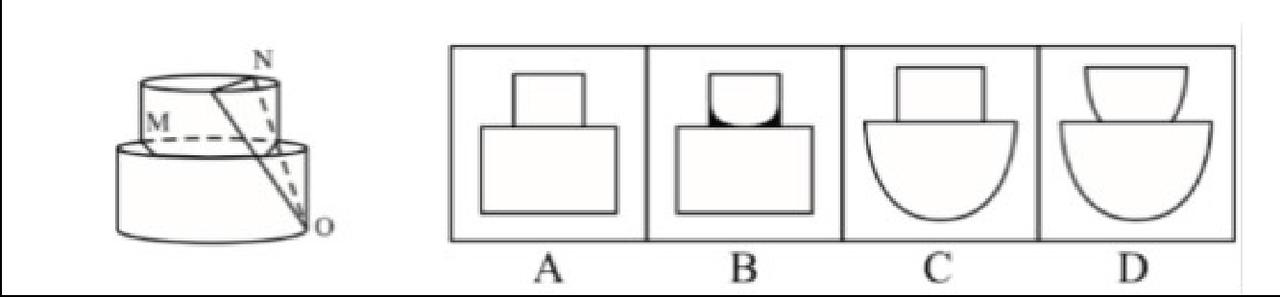} \\

\textbf{Q} & The cube below is obliquely cut along plane OMN; the section seen from the cutting plane may be: A B C D ? \\

\midrule
\textbf{A}  & D\\

\bottomrule
\end{tabular}
}
\vspace{2mm}
\caption{Example of Solid Cross-Section.}
\label{tab:visual_example_chichken}
\end{table*}

\begin{table*}[t]
\centering
\scalebox{0.88}{
\begin{tabular}{l p{0.9\textwidth}}
\toprule
\multicolumn{2}{l}{\bf 3D-Spatial Reasoning: Solid Assembly} \\
\midrule

& \includegraphics[width=0.6\textwidth,keepaspectratio]{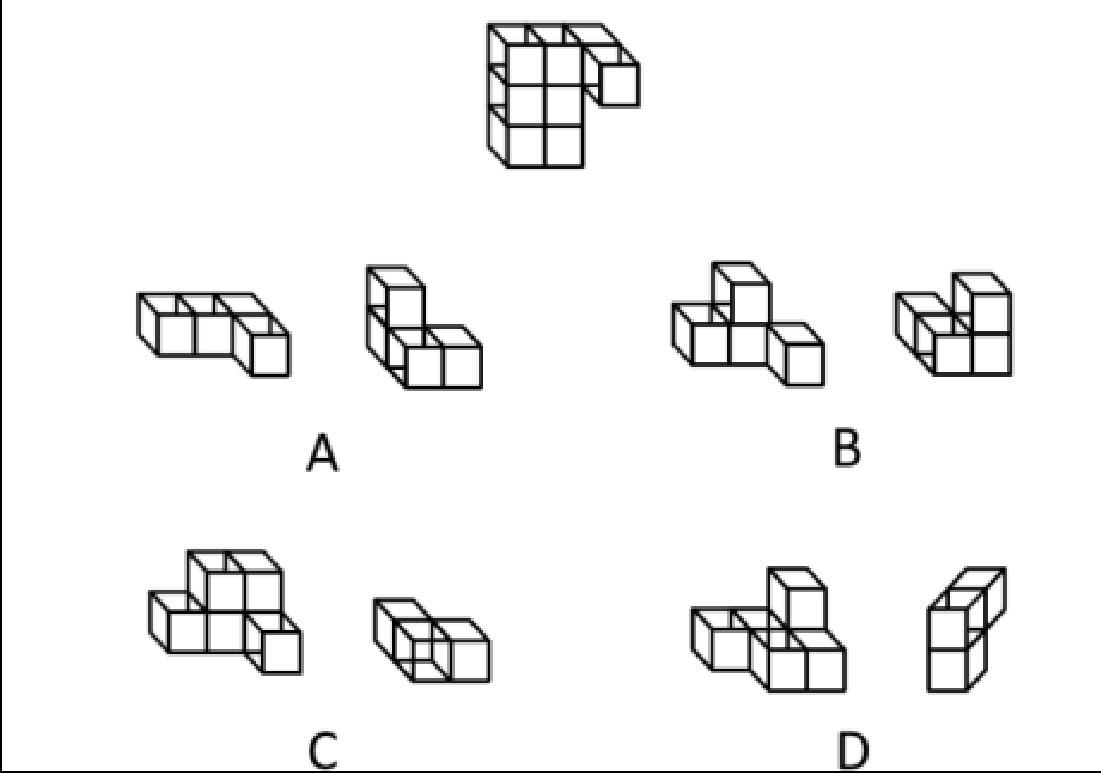} \\

\textbf{Q} & From the four given options, choose the most appropriate one to assemble the figure in the stem: A B C D ? \\

\midrule
\textbf{A}  & A\\

\bottomrule
\end{tabular}
}
\vspace{2mm}
\caption{Example of Solid Assembly.}
\label{tab:visual_example_chichken}
\end{table*}

\begin{table*}[t]
\centering
\scalebox{0.88}{
\begin{tabular}{l p{0.9\textwidth}}
\toprule
\multicolumn{2}{l}{\bf 3D-Spatial Reasoning: Three-view Drawing} \\
\midrule

& \includegraphics[width=0.6\textwidth,keepaspectratio]{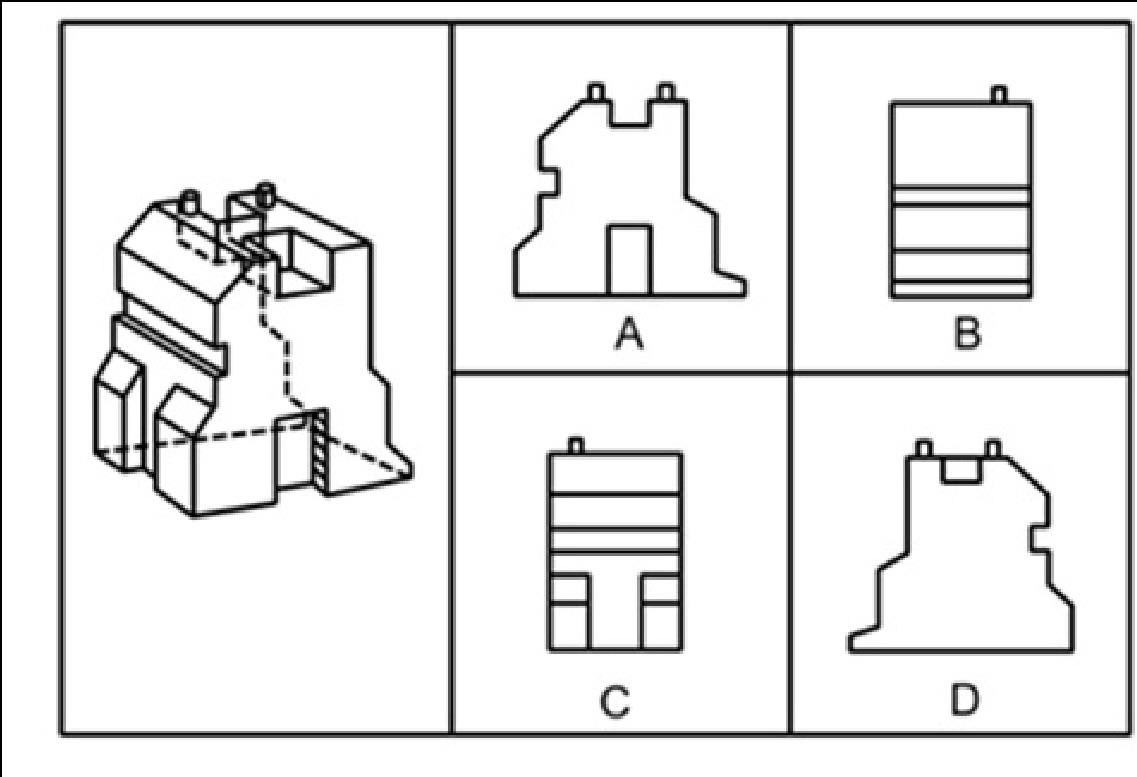} \\

\textbf{Q} & Among the four options on the right, which is not an elevation of the part on the left: A B C D ? \\

\midrule
\textbf{A}  & A\\

\bottomrule
\end{tabular}
}
\vspace{2mm}
\caption{Example of Three-view Drawing.}
\label{tab:visual_example_chichken}
\end{table*}

\begin{table*}[t]
\centering
\scalebox{0.88}{
\begin{tabular}{l p{0.9\textwidth}}
\toprule
\multicolumn{2}{l}{\bf Geolocation: Natural Landscapes} \\
\midrule

& \includegraphics[width=0.6\textwidth,keepaspectratio]{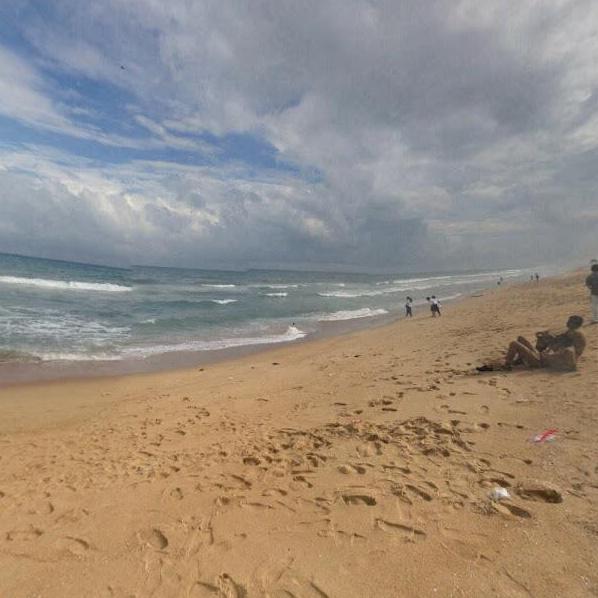} \\

\textbf{Q} & In which country and within which first-level administrative region of that country was this picture taken?Please answer in the format of country, administrative area level? \\

\midrule
\textbf{A}  & india,tamil nadu\\

\bottomrule
\end{tabular}
}
\vspace{2mm}
\caption{Example of Natural Landscapes.}
\label{tab:visual_example_chichken}
\end{table*}

\begin{table*}[t]
\centering
\scalebox{0.88}{
\begin{tabular}{l p{0.9\textwidth}}
\toprule
\multicolumn{2}{l}{\bf Geolocation: Architecture and Street Scenes} \\
\midrule

& \includegraphics[width=0.6\textwidth,keepaspectratio]{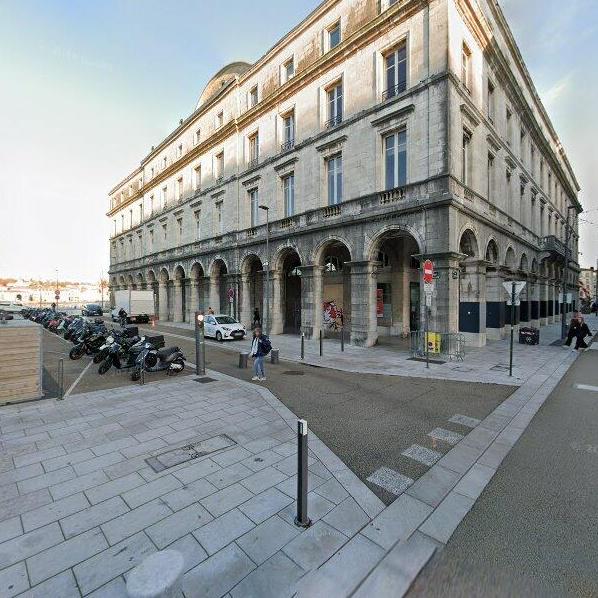} \\

\textbf{Q} & In which country and within which first-level administrative region of that country was this picture taken?Please answer in the format of country, administrative area level? \\

\midrule
\textbf{A}  & france, nouvelle-aquitaine \\

\bottomrule
\end{tabular}
}
\vspace{2mm}
\caption{Example of Architecture and Street Scenes.}
\label{tab:visual_example_chichken}
\end{table*}

\begin{table*}[t]
\centering
\scalebox{0.88}{
\begin{tabular}{l p{0.9\textwidth}}
\toprule
\multicolumn{2}{l}{\bf Geolocation: Indoor Scenes} \\
\midrule

& \includegraphics[width=0.6\textwidth,keepaspectratio]{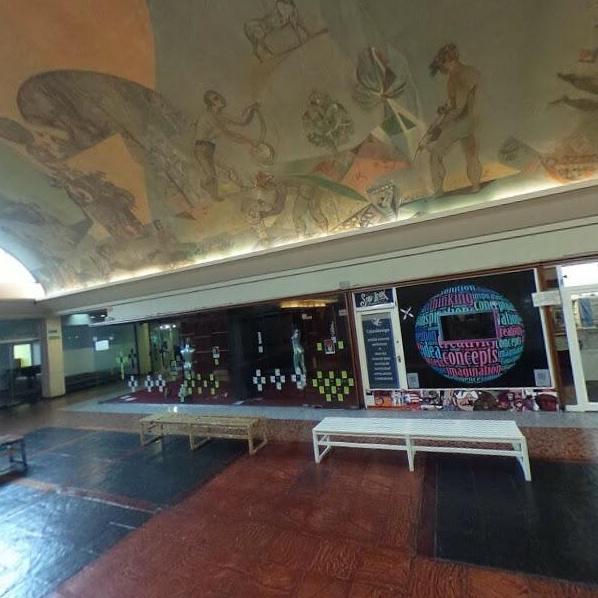} \\

\textbf{Q} & In which country and within which first-level administrative region of that country was this picture taken?Please answer in the format of country, administrative area level ?" \\

\midrule
\textbf{A}  & argentina, buenos aires\\

\bottomrule
\end{tabular}
}
\vspace{2mm}
\caption{Example of Indoor Scenes.}
\label{tab:visual_example_chichken}
\end{table*}

\begin{table*}[t]
\centering
\scalebox{0.88}{
\begin{tabular}{l p{0.9\textwidth}}
\toprule
\multicolumn{2}{l}{\bf Board Reasoning: Chess Position Evaluation} \\
\midrule

& \includegraphics[width=0.6\textwidth,keepaspectratio]{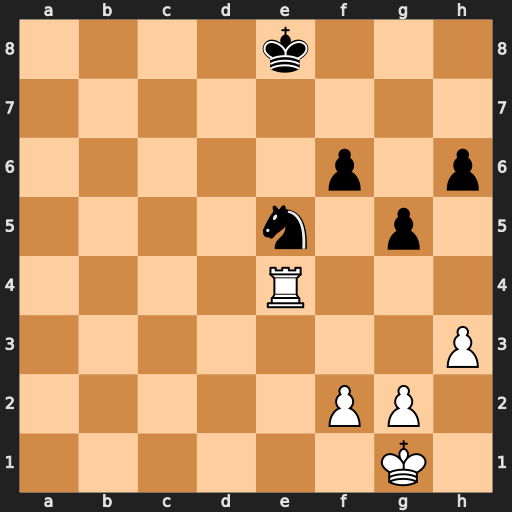} \\

\textbf{Q} & You are analyzing a chess position in FEN: 4k3/8/5p1p/4n1p1/4R3/7P/5PP1/6K1  w - - the Stockfish evaluation in centipawns (from White's perspective). Think deeper about this position: Don't just evaluate the current board state. Consider what the most likely moves are for both sides and how the centipawn evaluation would change as the position develops. Analyze a moves ahead - what does the future of this position look like? How would a strong engine assess this position after calculating many moves deep?Analyze step by step and explain your reasoning. \\

\midrule
\textbf{A}  & 400\\

\bottomrule
\end{tabular}
}
\vspace{2mm}
\caption{Example of Chess Position Evaluation.}
\label{tab:visual_example_chichken}
\end{table*}
\begin{table*}[t]
\centering
\scalebox{0.88}{
\begin{tabular}{l p{0.9\textwidth}}
\toprule
\multicolumn{2}{l}{\bf Board Reasoning: Chess Solution Sequence} \\
\midrule

& \includegraphics[width=0.6\textwidth,keepaspectratio]{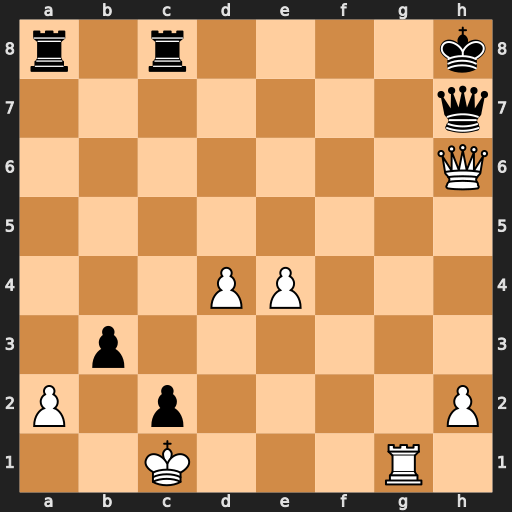} \\

\textbf{Q} & You are given a chess position in FEN: r1r4k/7q/7Q/8/3PP3/1p6/P1p4P/2K3R1 w - - 2 41.Find the best move for the side to play. \\

\midrule
\textbf{A}  & h6f6 \\

\bottomrule
\end{tabular}
}
\vspace{2mm}
\caption{Example of Chess Solution Sequence.}
\label{tab:visual_example_chichken}
\end{table*}

\begin{table*}[t]
\centering
\scalebox{0.88}{
\begin{tabular}{l p{0.9\textwidth}}
\toprule
\multicolumn{2}{l}{\bf Board Reasoning: Go Life-and-Death} \\
\midrule

& \includegraphics[width=0.4\textwidth,keepaspectratio]{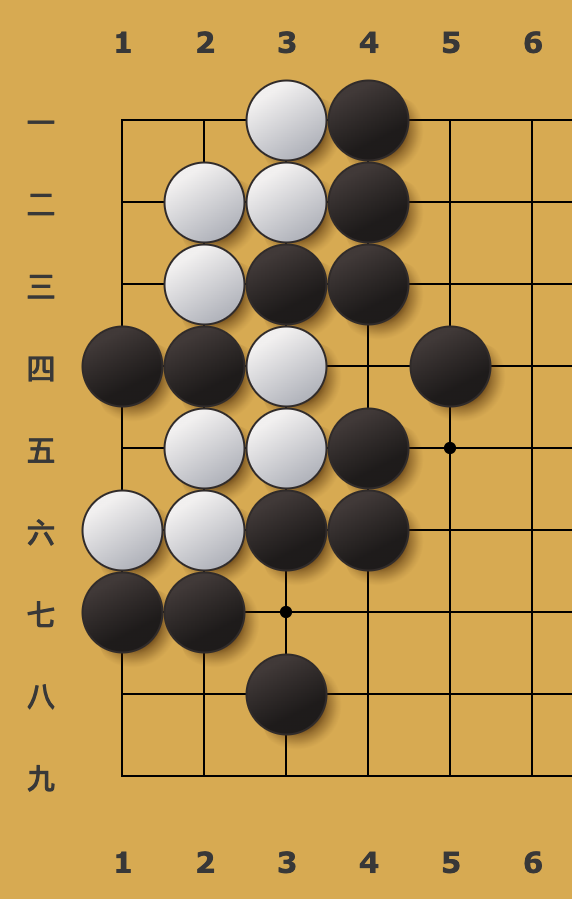} \\

\textbf{Q} & Go life \& death. Black to play. What's the single best move? Please answer with the coordinate. \\

\midrule
\textbf{A}  & \raisebox{0.25ex}{\rule{0.85em}{0.45pt}}2\\

\bottomrule
\end{tabular}
}
\vspace{2mm}
\caption{Example of Go Life-and-Death.}
\label{tab:visual_example_chichken}
\end{table*}

\begin{table*}[t]
\centering
\scalebox{0.88}{
\begin{tabular}{l p{0.9\textwidth}}
\toprule
\multicolumn{2}{l}{\bf Sudoku Solving: Standard 9 x 9 Sudoku} \\
\midrule

& \includegraphics[width=0.6\textwidth,keepaspectratio]{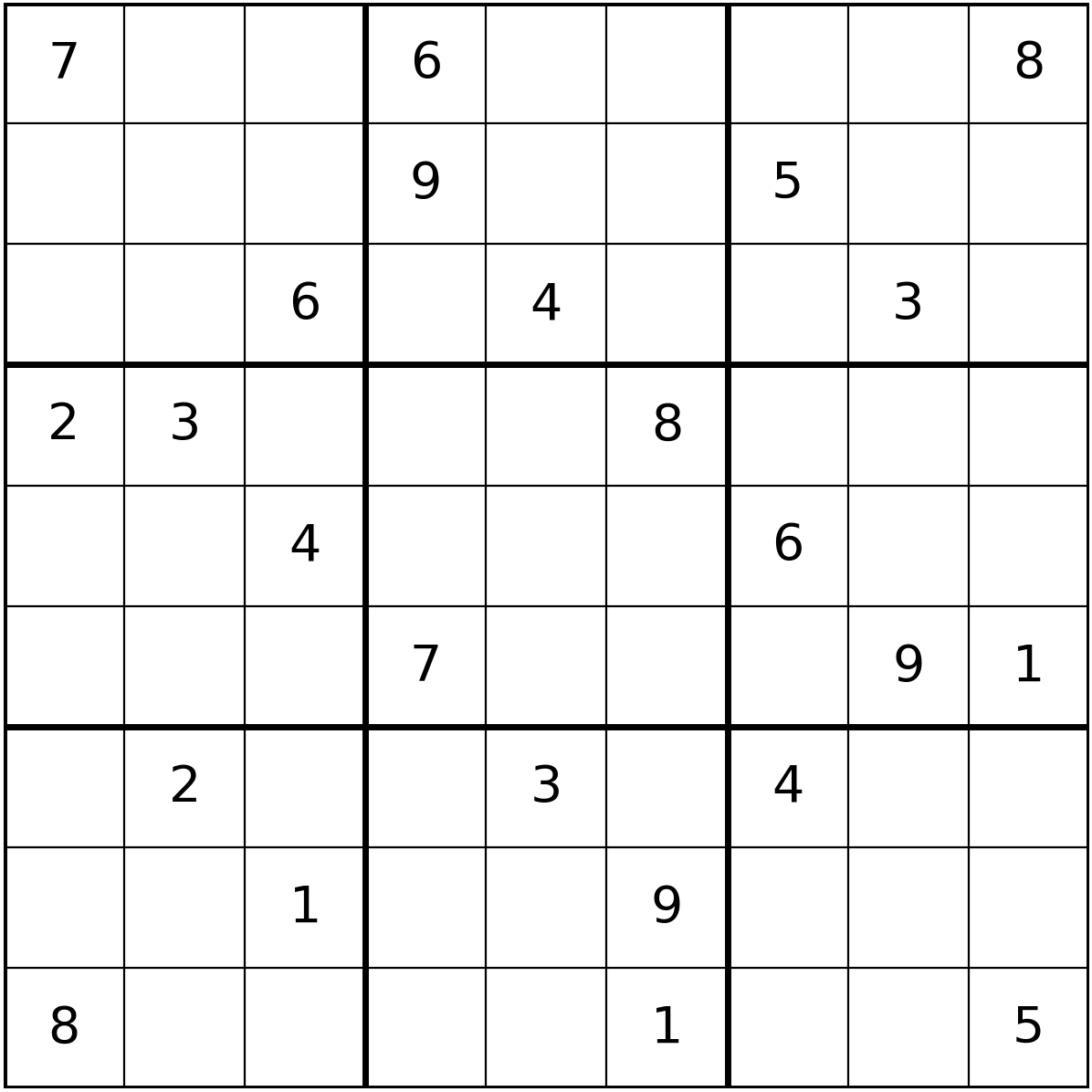} \\

\textbf{Q} & What is the digit at row 7, column 4 ? \\

\midrule
\textbf{A}  & 8\\

\bottomrule
\end{tabular}
}
\vspace{2mm}
\caption{Example of Standard 9 x 9 Sudoku.}
\label{tab:visual_example_chichken}
\end{table*}

\begin{table*}[t]
\centering
\scalebox{0.88}{
\begin{tabular}{l p{0.9\textwidth}}
\toprule
\multicolumn{2}{l}{\bf Cue Insight: Detective} \\
\midrule

& \includegraphics[width=0.6\textwidth,keepaspectratio]{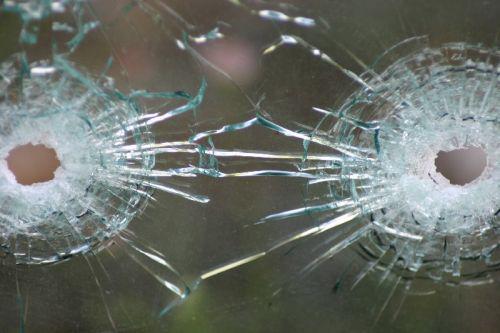} \\

\textbf{Q} & Two gunshots were fired through the window of a coffee shop. When the police arrived, they successfully recognized which gunshot was fired first. Which was the first gunshot and how did they figure that out? \\

\midrule
\textbf{A}  & The cracks of the left gunshot end up right at the cracks of the right gunshot. Therefore the first gunshot is the one on the right.\\

\bottomrule
\end{tabular}
}
\vspace{2mm}
\caption{Example of Detective.}
\label{tab:visual_example_chichken}
\end{table*}

\begin{table*}[t]
\centering
\scalebox{0.88}{
\begin{tabular}{l p{0.9\textwidth}}
\toprule
\multicolumn{2}{l}{\bf Cue Insight: Image Cue Extraction} \\
\midrule

& \includegraphics[width=0.6\textwidth,keepaspectratio]{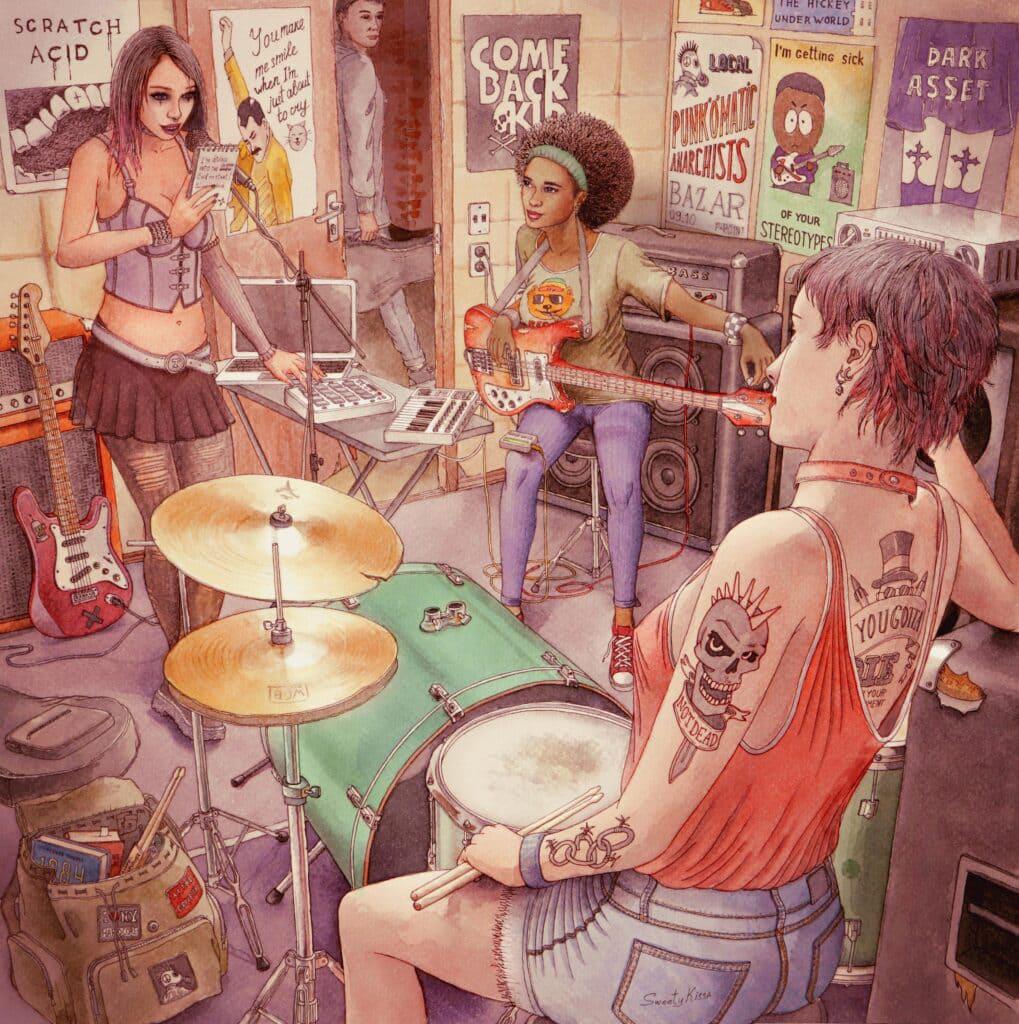} \\

\textbf{Q} & James sent this beautiful photo to a contest, hoping he will win first prize. The contest organizers, however, realized that the photo was fake and disqualified James. How did they figure out the picture was not real? \\

\midrule
\textbf{A}  & The vocalist’s name is Tessa. The bassist’s name is Delilah. The drummer’s name is Lisa.\\

\bottomrule
\end{tabular}
}
\vspace{2mm}
\caption{Example of Image Cue Extraction.}
\label{tab:visual_example_chichken}
\end{table*}

\begin{table*}[t]
\centering
\scalebox{0.88}{
\begin{tabular}{l p{0.9\textwidth}}
\toprule
\multicolumn{2}{l}{\bf Cue Insight: English Word Addition and Subtraction} \\
\midrule

& \includegraphics[width=0.6\textwidth,keepaspectratio]{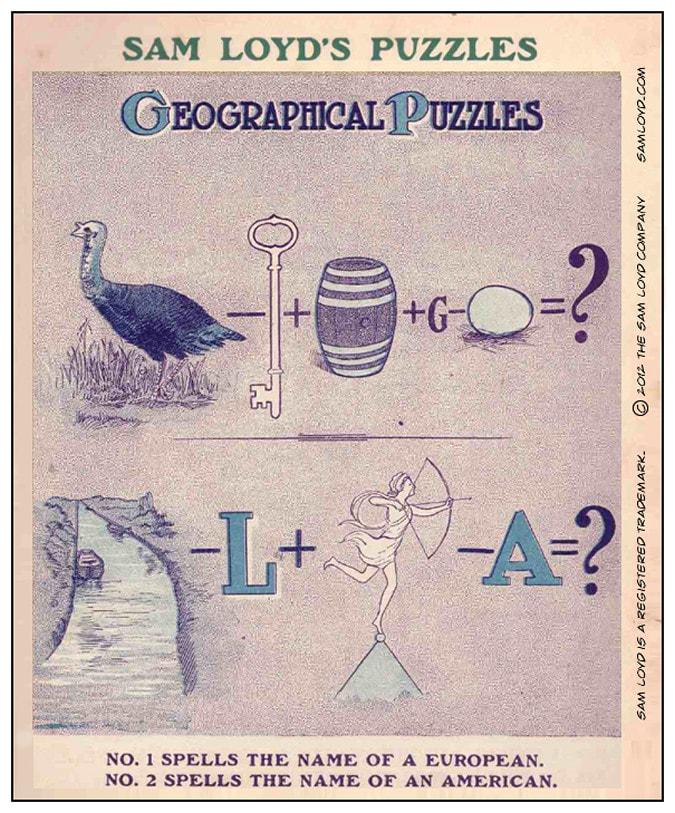} \\

\textbf{Q} & What nationalities do these sums spell? \\

\midrule
\textbf{A}  & CANADIAN\\

\bottomrule
\end{tabular}
}
\vspace{2mm}
\caption{Example of English Word Addition and Subtraction.}
\label{tab:visual_example_chichken}
\end{table*}

\begin{table*}[t]
\centering
\scalebox{0.88}{
\begin{tabular}{l p{0.9\textwidth}}
\toprule
\multicolumn{2}{l}{\bf Cue Insight: Folding-Perspective Insight} \\
\midrule

& \includegraphics[width=0.6\textwidth,keepaspectratio]{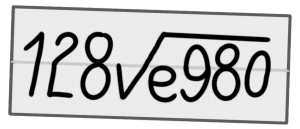} \\

\textbf{Q} & A boy receives a small note from a girl during his math class. He unfolds the note and sees the following expression. What does it mean?\\

\midrule
\textbf{A}  & If you fold the note the other way around, the message “I LOVE YOU” appears.\\

\bottomrule
\end{tabular}
}
\vspace{2mm}
\caption{Example of Folding-Perspective Insight.}
\label{tab:visual_example_chichken}
\end{table*}

\begin{table*}[t]
\centering
\scalebox{0.88}{
\begin{tabular}{l p{0.9\textwidth}}
\toprule
\multicolumn{2}{l}{\bf Cue Insight: Hidden-Word Decoding} \\
\midrule

& \includegraphics[width=0.6\textwidth,keepaspectratio]{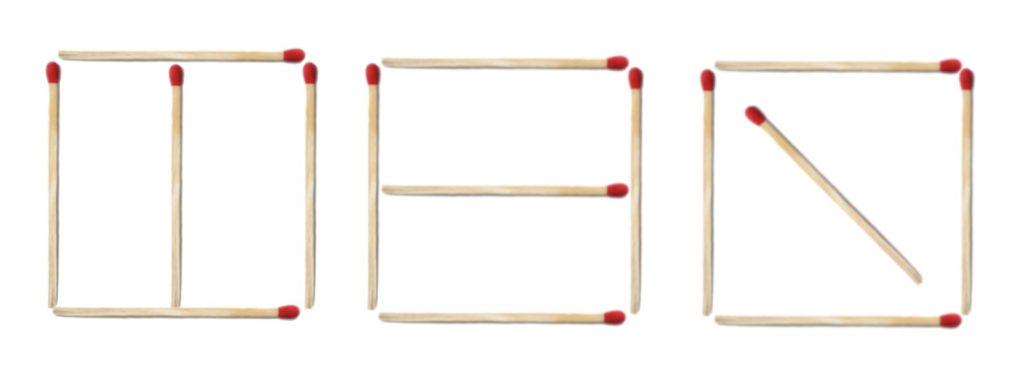} \\

\textbf{Q} & You have fifteen matchsticks. If you must remove exactly six of them, what word should the remaining matchsticks spell? \\

\midrule
\textbf{A}  & TEN\\

\bottomrule
\end{tabular}
}
\vspace{2mm}
\caption{Example of Hidden-Word Decoding.}
\label{tab:visual_example_chichken}
\end{table*}

\begin{table*}[t]
\centering
\scalebox{0.88}{
\begin{tabular}{l p{0.9\textwidth}}
\toprule
\multicolumn{2}{l}{\bf Cue Insight: Color-Set to Word} \\
\midrule

& \includegraphics[width=0.6\textwidth,keepaspectratio]{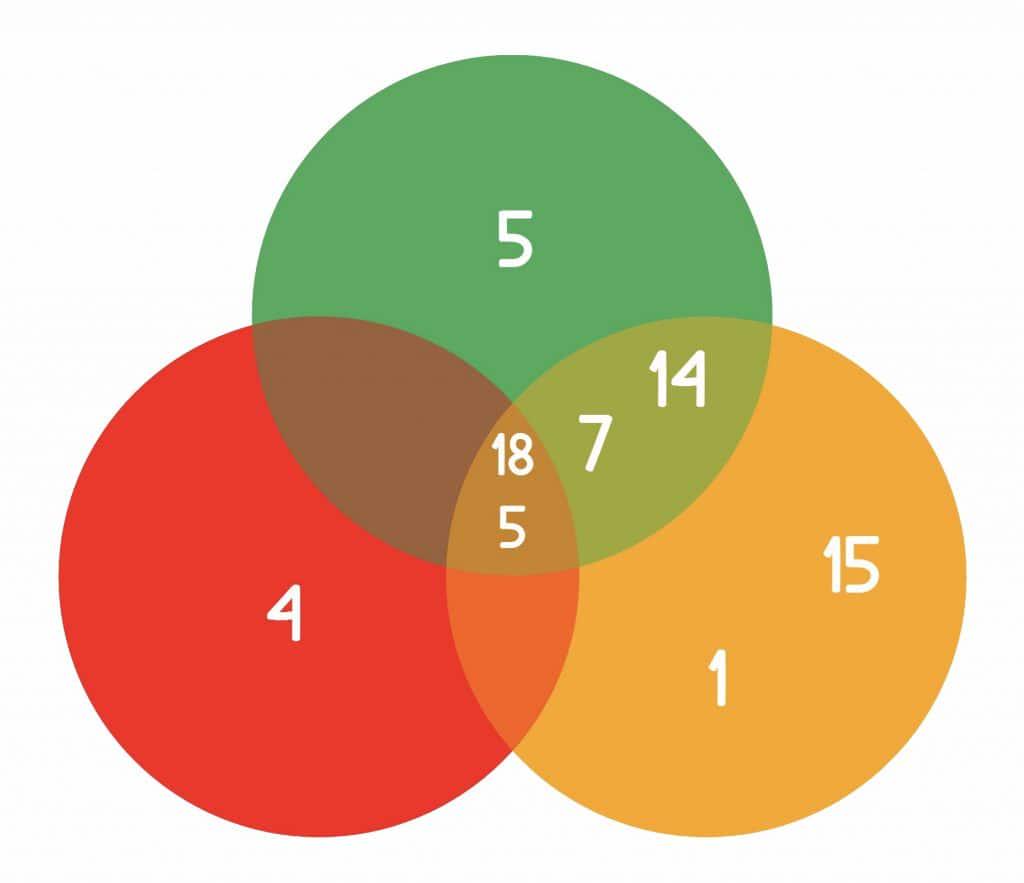} \\

\textbf{Q} & What does this Venn diagram depict? \\

\midrule
\textbf{A}  & Each of the numbers corresponds to a letter from the alphabet: 1 – A, 4 – D, 5 – E, 7 – G, 14 – N, 15 – O, 18 – R The three colors in the Venn diagram are GREEN, RED, ORANGE, and what is depicted is the letters they share.\\

\bottomrule
\end{tabular}
}
\vspace{2mm}
\caption{Example of Color-Set to Word.}
\label{tab:visual_example_chichken}
\end{table*}

\begin{table*}[t]
\centering
\scalebox{0.88}{
\begin{tabular}{l p{0.9\textwidth}}
\toprule
\multicolumn{2}{l}{\bf Inductive Reasoning: Rule Learning and Application} \\
\midrule

& \includegraphics[width=0.6\textwidth,keepaspectratio]{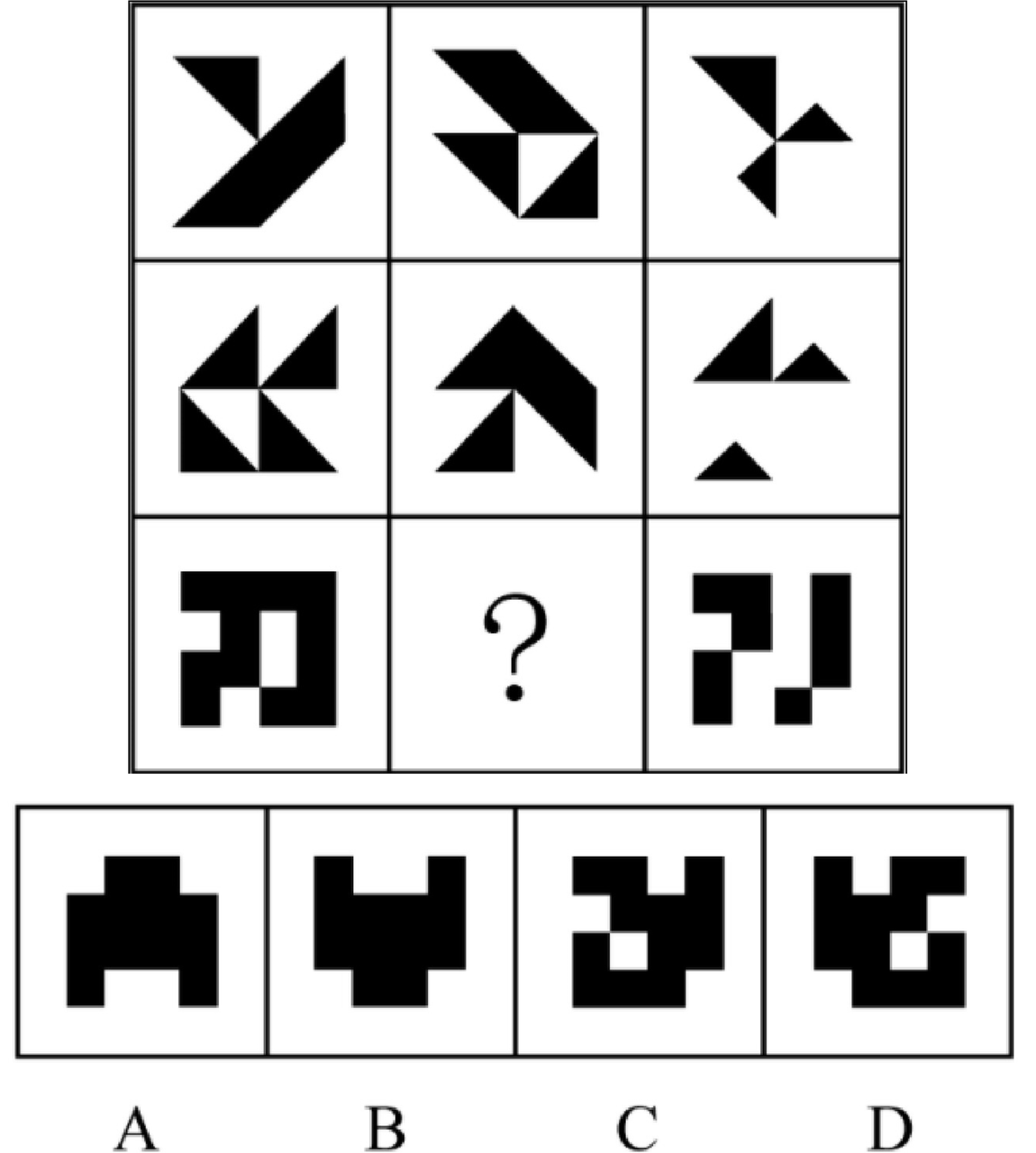} \\

\textbf{Q} & From the four options, choose the most appropriate one to fill in the question mark so that a pattern is presented: A B C D ? \\

\midrule
\textbf{A}  & C\\

\bottomrule
\end{tabular}
}
\vspace{2mm}
\caption{Example of Rule Learning and Application.}
\label{tab:visual_example_chichken}
\end{table*}

\begin{table*}[t]
\centering
\scalebox{0.88}{
\begin{tabular}{l p{0.9\textwidth}}
\toprule
\multicolumn{2}{l}{\bf Deductive Reasoning: Matrix Reasoning} \\
\midrule

& \includegraphics[width=0.6\textwidth,keepaspectratio]{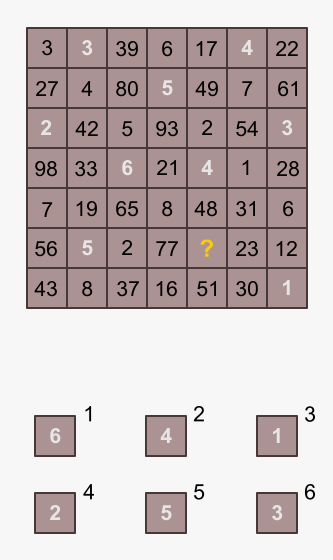} \\

\textbf{Q} & This difficult logic puzzle has 49 square cells with numbers, dark and light. The light numbers depend in some way on the dark numbers. You need to understand this relationship and choose the correct option for the light number in place of the question mark.\\

\midrule
\textbf{A}  & 2\\

\bottomrule
\end{tabular}
}
\vspace{2mm}
\caption{Example of Matrix Reasoning.}
\label{tab:visual_example_chichken}
\end{table*}

\begin{table*}[t]
\centering
\scalebox{0.88}{
\begin{tabular}{l p{0.9\textwidth}}
\toprule
\multicolumn{2}{l}{\bf Deductive Reasoning: Shape-Sequence Deductive Reasoning} \\
\midrule

& \includegraphics[width=0.6\textwidth,keepaspectratio]{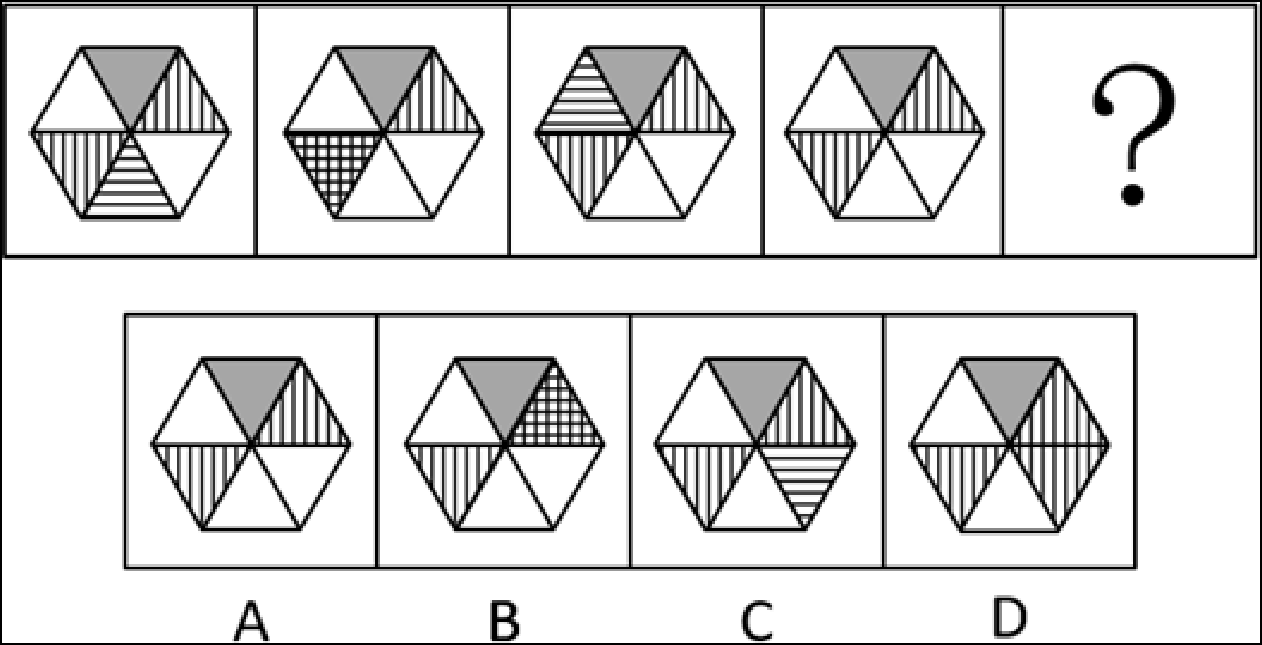} \\

\textbf{Q} &From the four options, choose the most appropriate one to fill in the question mark so that a pattern is presented: A B C D ? \\

\midrule
\textbf{A}  & B\\

\bottomrule
\end{tabular}
}
\vspace{2mm}
\caption{Example of Shape-Sequence Deductive Reasoning.}
\label{tab:visual_example_chichken}
\end{table*}

\end{document}